\documentclass[conference]{IEEEtran}

\usepackage{hyperref}
\usepackage{amsmath,amsfonts}
\usepackage{algorithmic}
\usepackage{algorithm}
\usepackage{array}
\usepackage[caption=false,font=normalsize,labelfont=sf,textfont=sf]{subfig}
\usepackage{textcomp}
\usepackage{stfloats}
\usepackage{url}
\usepackage{verbatim}
\usepackage{multirow}
\usepackage{graphicx}
\usepackage{cite}
\usepackage{xcolor}
\usepackage{bm}
\usepackage{amssymb}
\usepackage{booktabs}
\usepackage{capt-of}
\usepackage[most]{tcolorbox}
\definecolor{evblueBox}{HTML}{F0F7FF}
\usepackage[table]{xcolor}

\newcommand{\grayrow}{\rowcolor{gray!20}}
\renewcommand{\arraystretch}{1.5}  

\newif\ifdraft
\drafttrue

\ifdraft
    \newcommand{\todo}[1]{\textcolor{red}{\textbf{TODO: #1}}}
    \newcommand{\roberto}[1]{\textcolor{orange}{{[Roberto: #1]}}}
    \newcommand{\new}[1]{\textcolor{black}{{#1}}}
\else
    \newcommand{\todo}[1]{}
    \newcommand{\roberto}[1]{}
    \newcommand{\new}[1]{}
\fi

\usepackage{fancyhdr}
\begin{document}

\title{Motion-aware Event Suppression for Event Cameras}

\author{Roberto Pellerito, Nico Messikommer, Giovanni Cioffi, Marco Cannici, Davide Scaramuzza\\
Robotics and Perception Group, University of Zurich, Switzerland \\}





\twocolumn[{%
    \renewcommand\twocolumn[1][]{#1}%
    \maketitle
    \thispagestyle{fancy}  
    \vspace{-3.5ex}
    \begin{center}
    \includegraphics[width=\linewidth]{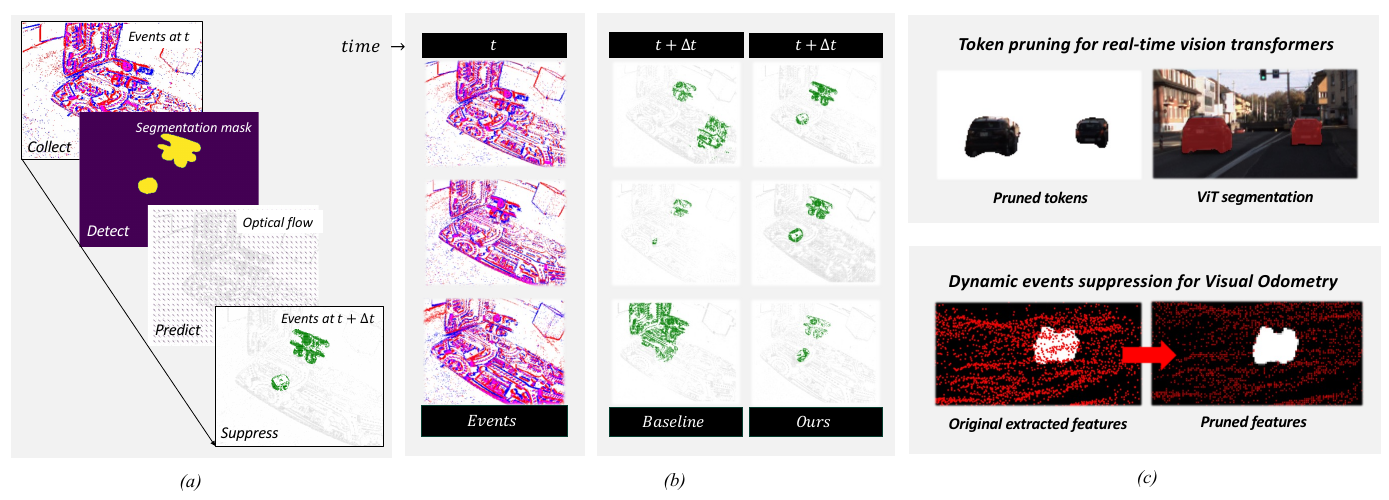}
    \captionof{figure}{%
    Our method disentangles ego-motion events from events triggered by independently moving objects (IMOs). (a) We jointly learn to segment IMOs and predict dense optical flow for the next $\Delta t$. Warping the mask forward yields an anticipated future mask, enabling suppression of future events. (b) Compared to baselines like EVIMO (baseline), our approach produces tighter masks with higher IoU and fewer false positives. (c) Event Suppression can be used for downstream tasks: (top) accelerating segmentation via motion-guided token pruning, and (bottom) improving visual odometry by filtering out dynamic IMO edges.
    }
  \label{fig:method_overview}
    \end{center}
    }]

\IEEEpeerreviewmaketitle

\let\thefootnote\relax\footnotetext{This work was supported by the European Union’s Horizon Europe Research and Innovation Programme under grant agreement No. 101120732 (AUTOASSESS), the European Research Council (ERC) under grant agreement No. 864042 (AGILEFLIGHT), and the Swiss AI Initiative by a grant from the Swiss National Supercomputing Centre (CSCS) under project ID a03 on Alps.}

\begin{abstract}
Event cameras report asynchronously per-pixel brightness changes with microsecond latency, encoding dynamic visual information as a sparse stream of events. However, their extreme temporal resolution floods perception systems with entangled events from ego-motion and independently moving objects (IMOs), which existing solutions fail to efficiently decouple, relying instead on prohibitive dense 3D reconstructions or limited hand-tuned filters. In this work, we introduce the first framework for Motion-aware Event Suppression, which learns to filter events triggered by IMOs and ego-motion in real time. 
Our model jointly segments IMOs in the current event stream while predicting their future motion, enabling anticipatory suppression of dynamic events before they occur.
Our lightweight architecture achieves 173 Hz inference on consumer-grade GPUs with less than 1 GB of memory usage, outperforming previous state-of-the-art methods on the challenging EVIMO benchmark by 67\% in segmentation accuracy while operating at a 53\% higher inference rate.
Moreover, we demonstrate significant benefits for downstream applications: our method accelerates Vision Transformer inference by 83\% via token pruning and improves event-based visual odometry accuracy, reducing Absolute Trajectory Error (ATE) by 13\%. 
\end{abstract}


\section*{Multimedia Material}
\textbf{Project Page:} \url{https://rpg.ifi.uzh.ch/event_suppression/} \\
\textbf{Code:} \url{https://github.com/uzh-rpg/event_suppression.git}

\section{Introduction}
\label{sec:Introduction}
 


Under constant illumination, events arise from two sources: background edges triggered by camera ego-motion and edges from independently moving objects. Because the sensor is uncapable of distinguishing natively between them, irrelevant events flood downstream perception pipelines, overloading computation and degrading accuracy. Our objective is to selectively suppress non-informative events while preserving those that matter for the task, an objective we term motion-aware event suppression.
 
Solving this problem is critical for emerging applications in AR/VR \cite{stoffregen2022event} and autonomous driving \cite{gehrig2024low}. For example, AR/VR systems require stable visual odometry (VO) to anchor virtual content while remaining responsive to gestures, while autonomous vehicles must track their own motion and react instantly to pedestrians. A reliable suppression mechanism is essential to ensure these systems can run at low latency while robustly perceiving relevant scene dynamics.

The challenge stems from an extreme data imbalance: as the camera moves, nearly every static object edge fires events, vastly outnumbering the few critical events triggered by moving objects. 
In realistic scenarios like \cite{gehrig2021dsec}, dynamic cues may constitute less than 5\% of the total event rate. 
This imbalance creates a chicken-and-egg dilemma: segmenting static background events requires knowing their motion context, yet observing this context comes only after those events have already flooded the system and introduced latency. Compounding this, each event carries minimal information (a pixel coordinate, timestamp, and polarity), making meaningful classification from single events impossible without additional context.

Previous approaches have largely focused on recovering intermediate representations such as camera poses, depth maps, or dense optical flow, which then feed into standard segmentation networks \cite{georgoulis2024out, wang2023evmoseg, zhang2023multi, mitrokhin2020learning}. Although these pipelines achieve solid benchmark performance, they suffer from high latency and inherit the fragility of complex SLAM systems, such as drift and depth estimation errors. Other methods using bio-inspired filtering \cite{parameshwara2021spikems, clerico2408retina} achieve low latency, but rely on manually tuned thresholds and have so far only been demonstrated indoors. Therefore the next frontier is a framework that can directly and adaptively anticipate and suppress motion in a task-driven, generalizable, and real-time manner.

The fundamental stumbling block is that existing methods either oversimplify by using static thresholds or overcomplicate by solving dense 3D scene reconstruction, which is unnecessary for suppression. They do not directly address the motion suppression challenge but instead attempt to solve auxiliary sub-problems, resulting in either brittle or computationally heavy pipelines. A principled, unified solution to separate ego-motion-induced events from independently moving object events, in real time, has been missing.

We introduce the first learning-based framework that unifies motion perception and prediction to address Motion-aware Event Suppression directly \new{(Fig. \ref{fig:method_overview})}. Our method is designed for practical deployment, achieving 173 Hz on a consumer-grade GPU while consuming less than 1 GB of memory, making it viable for latency-critical applications such as autonomous driving and AR/VR.

Our key idea is to predict future IMO and ego-motion events over a short horizon (up to 100 ms) by unifying instantaneous motion segmentation with motion flow forecasting. The segmentation head provides the necessary spatial context for reliable detection, while the flow prediction establishes a temporally grounded prior. This multi-task approach, combined with our time-conditioning module that learns a non-linear flow forecasting function from dense future supervision, enables precise anticipation of scene dynamics, facilitating real-time, low-latency event filtering.

Our framework sets a new state of the art on the challenging EVIMO \cite{mitrokhin2019ev} dataset, outperforming the previous best method by 67\% in segmentation accuracy while running 53\% faster. 
Furthermore we demonstrate practical benefits for downstream tasks: boosting Vision Transformer inference by 83\% FPS through motion-guided token pruning, and improving event-based VO accuracy by 13\% thanks to a 
high-signal event stream.

In summary, we introduce the first method to anticipate and decouple events from camera ego-motion and IMOs. 
Specifically:
\begin{itemize} 
\item By leveraging a multi-task architecture and training for predicting the future, our method establishes a new state of the art for both instantaneous and future IMO prediction \new{(Fig.~\ref{fig:method_overview}a)}. Using only event data, we outperform existing methods by 67\% in segmentation accuracy on classical IMOs segmentation benchmarks \new{(Fig.~\ref{fig:method_overview}b)}. 
\item We introduce a lightweight recurrent encoder-decoder featuring a novel cross-attention module (ATC). This design supports multi-horizon forecasting (Fig.~\ref{fig:time_ablation}) and achieves an inference rate of 173 Hz, surpassing the current state of the art by 53\% in inference frequency while maintaining a low memory footprint (Tab.~\ref{tab:run_time}). 
\item We demonstrate that anticipatory motion decoupling provides significant benefits to robotic perception \new{(Fig.~\ref{fig:method_overview}c)}, yielding an 11\% improvement in event-based visual odometry accuracy (Sec.~\ref{visualodometry}) and an 83\% acceleration in Vision Transformer (ViT) inference via motion-guided token pruning (Sec.~\ref{tokenpruning}). 
\end{itemize}

\section{Related Works}

To the best of our knowledge, this is the first work to introduce the task of event suppression, which aims to selectively ignore events triggered by dynamic or static objects based on their relevance to a given task. While prior research has extensively explored how to segment or track motion from event data, suppressing events themselves, either as a preprocessing step or as part of a downstream perception pipeline, has not been formally addressed. The most closely related line of research is event-based motion segmentation, where the goal is to identify and separate regions of the scene associated with independent motion. These methods typically aim to assign motion labels to pixels or regions rather than deciding whether events should be retained or discarded. Below, we review key developments in learning-based approaches to event-based motion segmentation.

\subsection{Learning-based motion segmentation using event cameras}
Deep neural networks have been widely applied to event-based motion segmentation. One of the earliest learning-based systems, Ev-DodgeNet \cite{sanket2020evdodgenet}, was developed for detecting and tracking moving objects in the context of dynamic obstacle avoidance. EV-IMO \cite{mitrokhin2019ev} introduced a structure-from-motion CNN adapted to event data for joint estimation of motion masks, depth, and camera motion, along with a dataset of indoor scenes featuring up to three independently moving objects.

While EV-IMO remains overall the state of art, subsequent methods improved different aspects of event-based motion segmentation. 
For instance, MSRNN \cite{zhang2023multi} proposed a multi-scale architecture to improve long-range temporal dependencies. GConv \cite{mitrokhin2020learning} introduces a Graph Neural Network, avoiding to construct frames and allowing for higher frequency of predictions.
Unsupervised methods have also emerged, such as Un-EVIMO \cite{wang2023evmoseg}, which learns from pseudo-labels generated as difference between estimated rigid and perceived optical flow.

These neural approaches often run in real-time and handle multiple moving objects, but early benchmarks focused on simplified indoor scenes. Recent work, such as \cite{georgoulis2024out}, extends motion segmentation to more complex indoor and outdoor settings using transformer-based spatial and temporal modules. However, such architectures typically demand high computational and memory resources, limiting their suitability for low-latency, real-time applications.

\subsection{Classical and Bio-Inspired Methods}
Early optimization-based methods leveraged event recency \cite{mitrokhin2018event} and energy minimization via spatio-temporal graph cuts \cite{zhou2021event} to segment moving objects. High-frequency estimates from these approaches were notably applied to autonomous flight for obstacle avoidance \cite{falanga2020dynamic}. Other works advanced per-event segmentation by clustering events based on motion hypotheses \cite{parameshwara20210} or utilizing iterative contrast maximization to align events and enhance sharpness \cite{stoffregen2019event, Gallego_2018_CVPR}. While these methods do not require training data, they often struggle under strong ego-motion and rely on iterative steps that limit real-time applicability in complex environments.

Alternatively, bio-inspired approaches employ neuromorphic principles for efficiency. SpikeMS \cite{parameshwara2021spikems} uses Spiking Neural Networks (SNNs) to handle temporal dynamics in continuous time. Other retina-inspired methods mimic biological Object Motion Sensitivity (OMS) through center-surround filtering \cite{clerico2408retina, d2025wandering}. Although computationally lightweight, these methods remain less accurate than deep learning models and require extensive parameter tuning.

\subsection{Low-latency Optical Flow Estimation from Event Data}

Optical flow estimation from event data is a closely related task to motion segmentation, as both aim to capture scene dynamics and object motion at high temporal resolution. A foundational work in this area, EV-FlowNet \cite{zhu2018ev}, introduced the first neural network to learn optical flow from events in an unsupervised manner by warping event data and minimizing a reconstruction loss. Building on this idea, \cite{ye2018unsupervised} proposed a joint framework to estimate dense optical flow, depth, and camera pose from event streams.

The release of larger datasets, such as DSEC \cite{gehrig2021dsec}, enabled the development and training of more powerful models like E-RAFT \cite{gehrig2021raft}, which achieved state-of-the-art performance in dense optical flow estimation for real-world driving scenarios. More recent works, including \cite{paredes2023taming} and \cite{shiba2022secrets}, have revisited unsupervised training from a contrast maximization perspective, leveraging the alignment of warped events to improve flow accuracy without ground-truth supervision.


\section{Methodology}

In this work, we address the task of Event Suppression by distinguishing between events caused by ego-motion and those originating from independently moving objects (IMOs). We begin by introducing the concept of Motion-Oriented Event Suppression and analyzing the impact of temporal latency on the estimation process in Section ~\ref{sec:staticeventsuppression}. Our approach anticipates future IMO motion, which we formalize in Section ~\ref{methodoverview}, and we implement it through a dedicated multi-task prediction network described in Section ~\ref{multitaskprediction}. To generate temporally-conditioned optical flow, our method employs a transformer-based architecture that incorporates future timestamps into its predictions, as detailed in Section ~\ref{timeconditioning}. Finally, the training process relies on multiple loss functions, which are discussed in Section ~\ref{losses}.

\subsection{Anticipatory Motion Suppression\label{sec:staticeventsuppression}}


We frame Anticipatory Motion Suppression as two coupled steps. First, \emph{event suppression}: every incoming event is labeled as coming either from an independently moving object or from camera ego-motion. This labeling relies on a binary mask that partitions the events into “retain” and “suppress” since we convert the event stream into an image-like grid.

Second, \emph{motion anticipation}: we predict how the scene will evolve over the next short time window. Concretely, we estimate a per-pixel motion field and warp the binary mask forward so that it aligns with the future events. \emph{Anticipatory Motion Suppression} is the composition of motion segmentation and successive motion prediction.



To formally introduce the task, we first define an event stream for a time interval of size $\Delta t$ as the set $\mathcal{E}$:
\begin{equation}
  \mathcal{E} 
  =\bigl\{e_i=(x_i,y_i,t_i,p_i)\bigr\}_{i=1}^{N},
  \quad p_i\!\in\!\{-1,+1\}
  \label{eq:event_set}
\end{equation}

Where $x_i,y_i$ represent the pixel location at which the event $e_i$ with polarity $p_i$ was triggered at timestep $t_i$.

\new{To simplify the notation, we define the set of events in the interval $[t-\Delta t, t]$ as $\mathcal{E}_{[t-\Delta t, t]} \equiv \mathcal{E} $.}

Let $\ell(e_i)\!\in\!\{0,1\}$ denote an oracle label that is
$1$ for events generated by independently moving objects (IMOs) and
$0$ for events caused solely by camera ego-motion.

Our goal is to find a mapping $\mathcal{S}$ defined for every event $e_i \in \mathcal{E}$ such that $\mathcal{S}$ maps all the $e_i$ whose true label $\ell(e_i) = 1$ to the partition of $\mathcal{E}$, which corresponds the set $\mathcal{E}'$ of all events associated with IMOs.\\
Such mapping $\mathcal{S}$ is described in Eq.\ref{eq:suppression_func}:

\begin{equation}
  \mathcal{S}:\mathcal{E}
  \longrightarrow
  \mathcal{E'}
  \label{eq:suppression_func}
\end{equation}

is called a motion-suppression operator \new{if and only if}
\begin{equation}
  \forall\,e\in\mathcal{E} \mid
  e\in\mathcal{S}(\mathcal{E})
  \iff
  \ell(e)=1 .
  \label{eq:ses_operator}
\end{equation}
The mapped set of event $\mathcal{E'}$ corresponds to the set of all IMO events $\mathcal{S}(\mathcal{E}) \equiv  \mathcal{E}'$.

The motion-suppression operator is defined continuously in time and can be applied asynchronously to each event. Note that $\mathcal{S}$ operates on a spatio\mbox{-}temporal window $\mathcal{E}$ and, in general, may itself be time-dependent within that window, since the IMO support at each instant $t'\!\in\![t-\Delta t,\,t)$ can vary significantly.

For tractability, we assume $\mathcal{S}$ is time-independent over $[t-\Delta t,\,t)$. Under this assumption, it suffices to determine the spatial support once (at a single reference time) and apply the same selection rule throughout the window.

Operationally, this corresponds to collapsing the time dimension of events into an image-like representation, estimating a mask $M_t$ with $\mathcal{S}$, and then using $M_t$ to retain the $(x,y)$ locations associated with IMOs when gating events.

Let
\(
  M_t \!\in\!\{0,1\}^{H\times W}
\)
denote the Pixel-wise instantaneous IMO mask for time interval $(t-\Delta t, t)$:
\[
  M_t(x,y)=
  \begin{cases}
    1 &\text{if the pixel $(x,y)$ belongs to an IMO at $t$ },\\[2pt]
    0 &\text{otherwise.}
  \end{cases}
\]
Under this view, the motion-suppression operator introduced in
\eqref{eq:ses_operator} can be rewritten as
\begin{equation}
  \mathcal{S}_{M_t}\bigl(\mathcal{E} \bigr)
  \;=\;
  \Bigl\{\,e_i \,\big|\, M_t(x_i,y_i)=1\Bigr\},
  \label{eq:pixel_operator}
\end{equation}
i.e.\ every event is retained  iff it is triggered by a pixel currently
marked dynamic.

Naturally, each algorithm predicting dynamic object masks from an event slice
\(
  \mathcal{E} 
\)
features a processing time $\Delta t_d$, which leads to a lagging estimate
\(
  \widehat{M}_{t-\Delta t_d}
\)
of the pixel-wise dynamic-object mask.
Using this lagging estimate introduces falsely predicted dynamic masks
\begin{equation}
  \mathcal{S}_{\widehat{M}_{t-\Delta t}}
    \bigl(\mathcal{E} \bigr)
  \;\neq\;
  \mathcal{S}_{M_t}
  \bigl(\mathcal{E} \bigr),
  \label{eq:delyed_suppression}
\end{equation}
whenever moving objects have translated by more than one pixel during $\Delta t$.
This temporal latency is the root cause of the misalignment between real and predicted dynamic-events.

To solve this misalingment, we propose to warp the predicted mask into the future using a dense flow field
\(
  \psi_{t+\Delta t}\!\in\!\mathbb{R}^{2\times H\times W}
\)
that predicts the pixel motion for $\,\Delta t$ ms into the future.  
Using the future flow field and the lagging estimate $\widehat{M}$, we obtain an estimate $\widetilde{M}$ of the pixel-wise instantaneous dynamic object mask 
\begin{equation}
  \widetilde{M}_t(\mathbf{x})
  \;=\;
  \widehat{M}_{t-\Delta t}
  \bigl(\mathbf{x}-\psi_t(\mathbf{x})\bigr),
  \qquad
  \mathbf{x}=(x,y),
  \label{eq:warp_mask}
\end{equation}


Because \(\widetilde{M}_t\) is spatially aligned with the current
scene, the operator in \eqref{eq:pixel_operator} \new{with $M_t=\widetilde{M}_t$} realizes anticipatory motion suppression even though it relies
only on past observations.

\begin{figure*}[t] 
  \centering
  \includegraphics[width=0.83\textwidth]{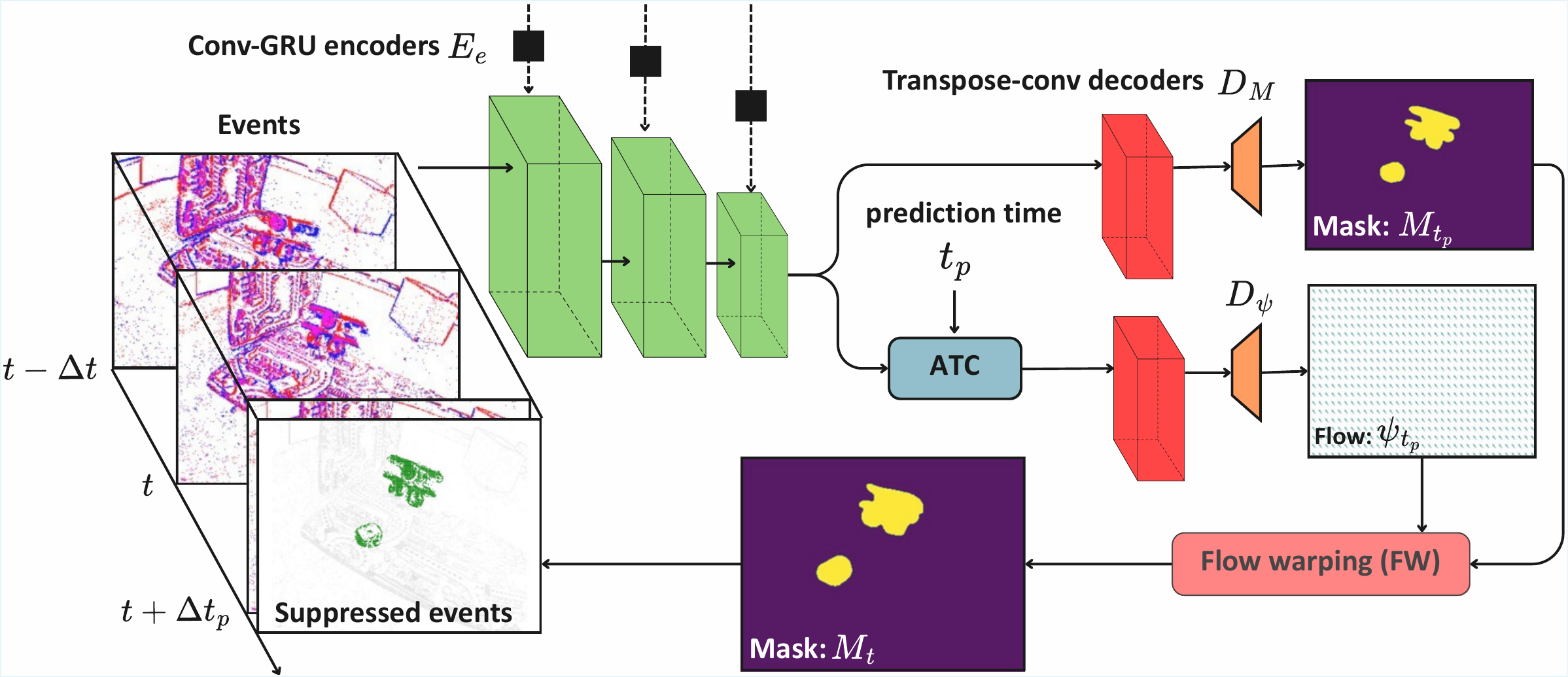}
    \captionof{figure}{%
    \textbf{Overview of the Anticipatory Motion Suppression pipeline} A stack of input events $\mathcal{E} $ is processed by a network featuring a series of recurrent blocks $E_e$ followed by our proposed attention-based time conditioning (ATC). \new{Raw features from events are fed to the transposed convolutional decoders $D_{M}$ for IMO mask prediction}. \new{Time conditioned features are used by a second transposed convolutional decoder $D_{\psi}$ for flow forecast prediction}. The network finally predicts a binary dynamic-object mask $M_t$ and a future dense optical flow $\psi_{t \rightarrow t + \Delta t_p}$.
    The predicted flow is used in a flow warping module (FW) to forecast the motion of dynamic regions.
    By propagating the mask forward in time, the system anticipates and suppresses future events $\mathcal{E}_{[t, t+\Delta t_p)}$ corresponding to either independently moving objects or the static background, yielding a simplified event stream focused on just one specific type of motion.
    }
  \label{fig:method_overview_real}
\end{figure*}

\subsection{Method Overview\label{methodoverview}}
As depicted in Fig. \ref{fig:method_overview_real}, our model operates on events $\mathcal{E} $ from a $\Delta t$ temporal interval, which are first converted into a dense $B\times H\times W$ stack representation \cite{mostafavi2021learning} with a temporal discretization of 2 bins. To achieve a unified understanding of scene dynamics, the network then performs two tasks simultaneously. First, it generates a binary mask $M_t$ at spatial resolution $ H \times W$, segmenting events caused by independently moving objects from the events triggered by the static background. 

This binary mask \(M_t \!\in\!\{0,1\}^{H\times W}\) classifies pixels as 0 if they are associated with events generated by static objects and as 1 if the corresponding events are triggered by an IMO. 
As a second task, guided by an input forecast time $t_p$, the network predicts the future dense optical flow $\psi_{t \rightarrow t+t_p}$ to forecast scene motion

\subsection{Multi-Task Prediction\label{multitaskprediction}}



Inspired by \cite{zhu2018ev, Paredes-Valles_2023_ICCV}, we adopt an $n$-stage conv-GRU encoder~\cite{ballas2015delving}, selected for its efficiency and suitability for real-time embedded applications. The features from this encoder are fed into two parallel branches. In the first branch, the raw features are passed to a mask-specific decoder to predict the current motion segmentation mask $M_t$. In the second, the features are processed by the ATC module (Fig.~\ref{fig:method_overview_real}) to generate time-conditioned features, which are then fed to the optical flow decoder to predict future flow $\psi_{t \rightarrow t+\Delta t_p}$. Both decoders consist of $n$ transposed convolutional layers. 

Overall, this framework relies on two key components: Attention-based Time Conditioning and Flow Warping.
\begin{figure}[t] 
  \centering
  \includegraphics[width=\columnwidth]{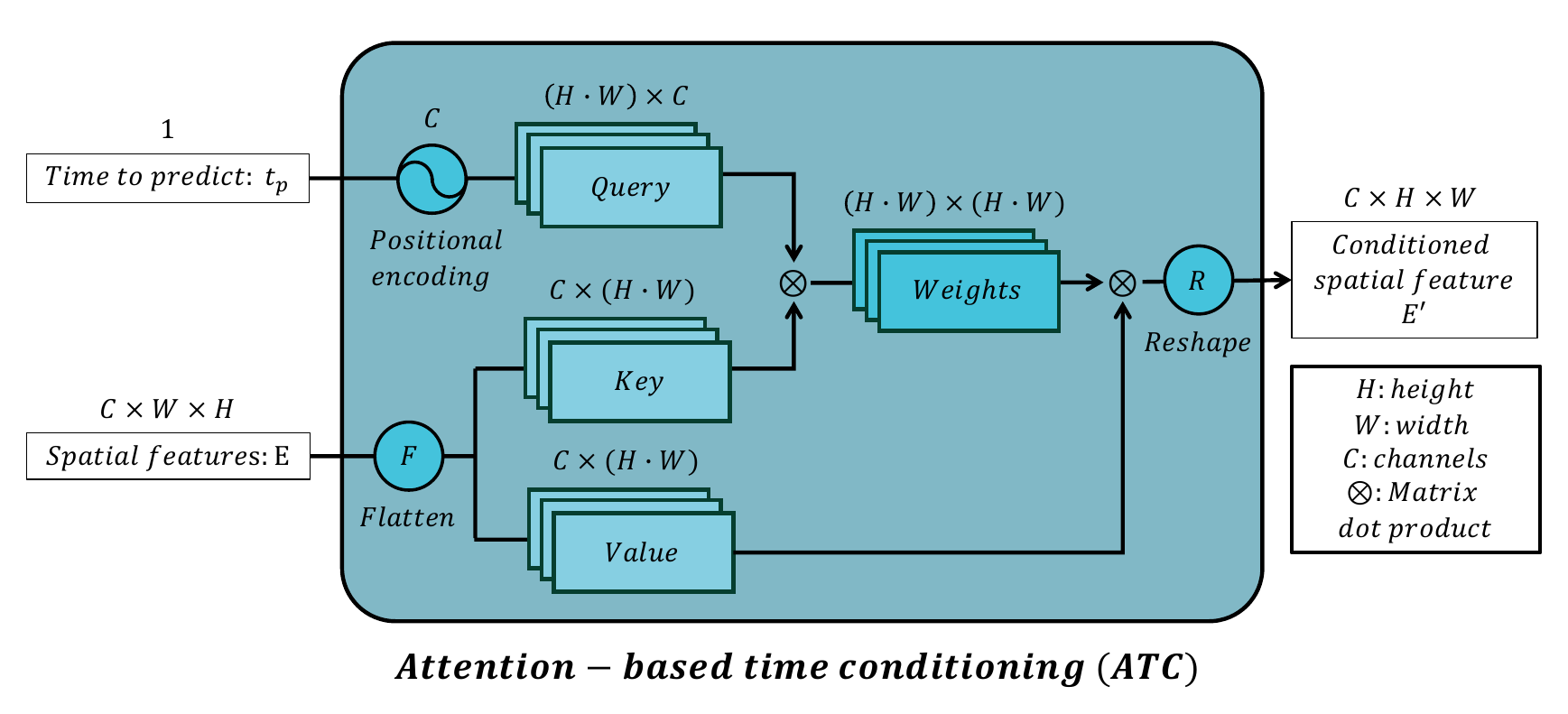}
  \captionof{figure}{%
 \textbf{Overview of Attention-based Time Conditioning (ATC).} Target time $\Delta t_p$ is mapped via Positional Encoding (PE) to an embedding of size $C$, matching the spatial feature channels. These temporal features are broadcasted to form a Query, while flattened spatial features serve as the Key and Value. Cross-attention modulates the spatial features based on the temporal query, yielding the time-conditioned embedding $E'$.
  }
  \label{fig:atc}
\end{figure}



\subsubsection{Attention-based Time Conditioning (ATC)\label{timeconditioning}}
To forecast optical flow at an arbitrary future time $t + \Delta t_p$, we introduce an ATC module that modulates the current spatial feature embedding $\mathbf{E} \in \mathbb{R}^{B \times C \times H \times W}$ based on the target time delta. As illustrated in Fig.~\ref{fig:atc}, we employ a multi-head cross-attention mechanism where the Query $\mathbf{Q}$ is derived from the temporal positional encoding $\mathbf{t}_{\text{enc}} = \text{PE}(\Delta t)$ broadcasted across all spatial positions. The Key $\mathbf{K}$ and Value $\mathbf{V}$ are obtained by flattening the spatial dimensions of $\mathbf{E}$ into a sequence. The cross-attention operation produces a modulated sequence $\mathbf{E}'$ that is reshaped back to the original spatial dimensions, yielding the time-conditioned embedding $\mathbf{E}_{t+\Delta t}$ for the optical flow decoder. Further details on the attention input preparation, sequence formatting and positional encoding are provided in the Appendix (Sec.~\ref{sec:appendix_positionalencoding}).

\subsubsection{Mask and Flow Decoding \label{sec:decoding}}
The features coming from the ATC module $\mathbf{E}_{t+\Delta t}$ and the event encoder $\mathbf{E}_t$ are further decoded to obtain an optical flow of the scene and a binary mask of the IMOs.
The decoders for optical flow $D_{\psi}$ and binary mask prediction $D_{M}$, depicted in Fig.~~\ref{fig:method_overview_real}, consist of a series of transposed convolutional layers. 
Since each data modality requires a specific feature representation, the two decoders have two distinct sets of weights. To retain spatial information for both the encoded flow $\mathbf{E}_{t+\Delta t}$ and the mask $\mathbf{E}$, we add skip connections from the encoding blocks of the encoder to the decoders.

\subsubsection{Mask flow warping \label{sec:mask_flow_warping}}
We combine the outputs of the  Mask and  Flow decoders to predict the future dynamic-object segmentation. Concretely, we warp the  mask logits from time $t$ to $t{+}1$ using  backward warping with the forward flow $\psi_{t_p}$ as expressed in Equation~\ref{eq:warp_mask}, which is implemented via differentiable sampling with bilinear interpolation. We propagate soft (logit) masks rather than hard labels and train with robust losses e.g. Charbonnier \cite{charbonnier1997deterministic}, which naturally down-weights residual outliers near independently moving object (IMO) boundaries and near occluded regions.

\subsection{Training Losses\label{losses}}
Our model is trained end-to-end using a hybrid loss $\mathcal{L}_{\text{total}} = w_{\text{sup}}\mathcal{L}_{\text{sup}} + w_{\text{unsup}}\mathcal{L}_{\text{unsup}}$. The supervised term $\mathcal{L}_{\text{sup}}$ combines a Binary Cross-Entropy and Dice loss \cite{li2019dice} for IMO segmentation at current ($M_t$) and future ($M_{t+\Delta t}$) timestamps, alongside an $L_1$ and Charbonnier smoothness loss for optical flow\cite{charbonnier1997deterministic}. To leverage raw event data without labels, $\mathcal{L}_{\text{unsup}}$ employs a contrast maximization objective insipired by~\cite{Gallego_2019_CVPR, Paredes-Valles_2023_ICCV} . This focus metric encourages flow fields that "deblur" the event stream by iteratively warping events to a reference time $t_{ref}$ to maximize sharpness. Further details on loss formulations, weighting factors, and hyperparameter values are provided in the Appendix (Sec.~\ref{sec:appendix_losses}).

\section{Experiments}

\begin{figure}[t]
    \centering
    \includegraphics[width=\columnwidth]{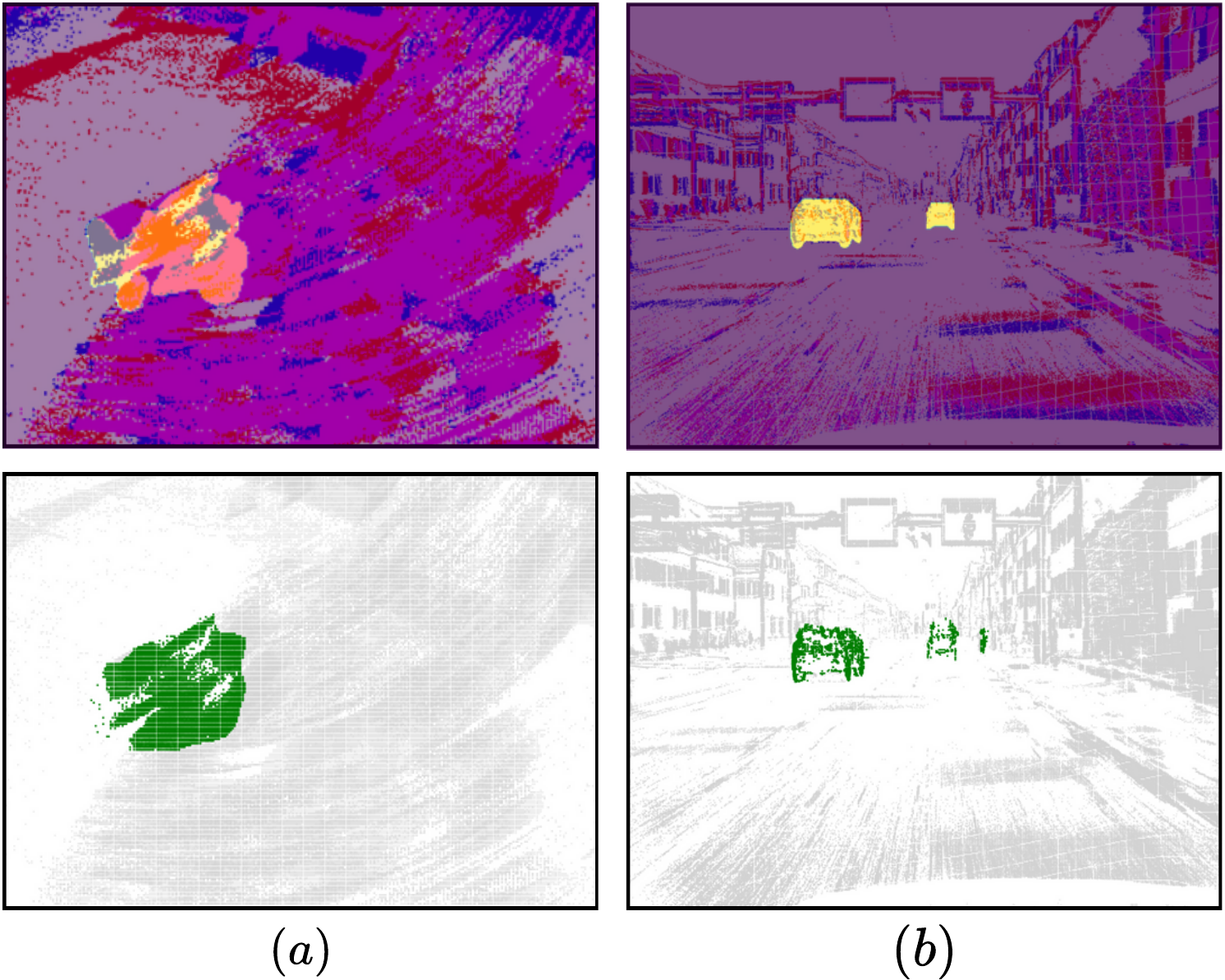}
    \captionof{figure}{
        Anticipatory motion suppression on EVIMO (a) and DSEC (b). Top row: Events accumulated over $\Delta t_p = 100$\,ms with ground-truth IMO masks. Bottom row: Predicted separation of ego-motion (white) and IMO (green) events for the future 100\,ms window based on the previous 50\,ms of data. Our pipeline demonstrates: (a) robustness under extreme ego-motion with complex IMO shapes and (b) accurate motion anticipation for both distant and nearby vehicles in driving scenes.
        }
    \label{fig:model_results}
\end{figure}

We evaluate our framework on the EVIMO \cite{mitrokhin2019ev}, DSEC \cite{gehrig2021dsec}, and EED \cite{EED_mitrokhin2018event} datasets, with qualitative results shown in \new{Fig.~\ref{fig:model_results}}. We first establish state-of-the-art performance in future motion forecasting (Sec.~\ref{sec:futureevimotest}) and instantaneous motion segmentation on EVIMO against a comprehensive suite of supervised, unsupervised, and frame-based baselines (Sec.~\ref{sec:evimotest}). We then demonstrate our model's robustness to low illumination and severe sensor noise by benchmarking on the challenging EED dataset (Sec.~\ref{sec:EEDtest}). Furthermore, we analyze the system's computational efficiency, highlighting favorable trade-offs between runtime, prediction age, and GFLOPS (Sec.~\ref{timecomparison}). The necessity of our core architectural components, model capacity scaling, and performance sensitivity across varying prediction horizons are systematically validated through extensive ablation studies (Sec.~\ref{ablations}, \ref{sec:pred_horizon_ablation}). Finally, we showcase the practical utility of our anticipatory suppression mechanism in two downstream robotic tasks: improving event-based visual odometry accuracy (Sec.~\ref{visualodometry}) and accelerating Vision Transformer inference via motion-guided token pruning (Sec.~\ref{tokenpruning}).

\subsection{Training Setup \label{sec:train_setup}}
We train our model on the EVIMO \cite{mitrokhin2019ev} and DSEC \cite{gehrig2021dsec} training splits, aggregating events into 50\,ms windows encoded as two-bin voxel grids. Since ground truth is provided at 10\,Hz, we supervise the mask loss $\mathcal{L}_{\text{mask}}$ at 100\,ms intervals and compute $\mathcal{L}_{\text{unsup}}$ from the corresponding event windows. For EVIMO, which lacks ground-truth flow, we disable $\mathcal{L}_{\text{flow\_sup}}$. On DSEC, we apply $\mathcal{L}_{\text{flow\_sup}}$ only to static regions, as the provided flow doesn't take in account IMOs. Detailed loss weights, optimization parameters, and implementation details are provided in the Appendix (Sec.~\ref{sec:appendix_train_setup}).

\subsection{Future Motion Segmentation on EVIMO \label{sec:futureevimotest}}
\begin{table}[t]
\footnotesize
\renewcommand{\arraystretch}{1.2}
\caption{Mean Intersection Over Union \% (mIoU) and ratio of correctly segmented objects \% (R@0.5) for future prediction on EVIMO dataset.}
\centering
\setlength{\tabcolsep}{3pt}
\begin{tabular}{l cccc cccc}
\toprule
 & \multicolumn{2}{c}{\textbf{OMS} \cite{clerico2408retina}} & \multicolumn{2}{c}{\textbf{EV-IMO$^\dagger$} \cite{mitrokhin2019ev}} & \multicolumn{2}{c}{\textbf{EV-IMO$^*$} \cite{mitrokhin2019ev}} & \multicolumn{2}{c}{\textbf{Ours}} \\
\cmidrule(lr){2-3} \cmidrule(lr){4-5} \cmidrule(lr){6-7} \cmidrule(lr){8-9}
\textbf{Sequence} & \textbf{mIoU} & \textbf{R@0.5} & \textbf{mIoU} & \textbf{R@0.5} & \textbf{mIoU} & \textbf{R@0.5} & \textbf{mIoU} & \textbf{R@0.5} \\
\midrule
Boxes & 49.15 & 2.21 & 69.74 & 53.73 & \underline{73.53} & \underline{61.05} & \textbf{76.24} & \textbf{75.38} \\
\rowcolor[gray]{0.9}
Floor & 48.23 & 0 & \underline{77.06} & 55.95 & 75.47 & \underline{66.93} & \textbf{80.10} & \textbf{95.51} \\
Wall & 48.63 & 0.27 & 75.45 & 40.02 & \underline{75.78} & \underline{68.06} & \textbf{79.63} & \textbf{88.09} \\
\rowcolor[gray]{0.9}
Table & 48.73 & 0.79 & \underline{78.41} & 48.37 & 75.26 & \underline{61.23} & \textbf{79.71} & \textbf{90.95} \\
Fast & 45.97 & 0.69 & 63.96 & 33.03 & \underline{64.66} & \underline{38.94} & \textbf{68.07} & \textbf{67.26} \\
\rowcolor[gray]{0.9}
Tabletop & 50.77 & 0.90 & \underline{78.51} & 73.44 & 72.51 & \underline{54.46} & \textbf{84.57} & \textbf{91.48} \\
\bottomrule
\end{tabular}
\label{tab:evimo_future_pred}
\end{table}
We evaluate 100\,ms motion forecasting against EVIMO~\cite{mitrokhin2019ev} and OMS~\cite{clerico2408retina} (Tab.~\ref{tab:evimo_future_pred}). We retrain EVIMO for future prediction using future depth supervision. Since the default EVIMO architecture processes a symmetric window (accessing 50\,ms of future data), we report results for both the standard non-causal version (\new{namely} EVIMO$^*$) \new{using past and future frames} and a strict causal variant (\new{namely} EVIMO$^\dagger$) restricted to past frames. OMS is evaluated using optimal parameters as a lightweight baseline. Performance is measured via mIoU and R@0.5 against ground-truth masks shifted 100\,ms forward. Detailed metric definitions are provided in Appendix~\ref{sec:appendix_metrics}.

As shown in Tab.~\ref{tab:evimo_future_pred}, our simple pipeline consistently outperforms EVIMO$^*$ and EVIMO$^\dagger$ on the instance-level metric $R@0.5$ by 45\% and 67\%, respectively, despite requiring neither future frames at run time nor ground-truth depths during training. 
On mIoU, our method also surpasses EVIMO$^*$ and EVIMO$^\dagger$ by 7\% and 6\%, respectively, demonstrating strong segmentation of both IMO and ego-motion events. In contrast, OMS struggles to capture true IMOs, producing many false positives and achieving an average $R@0.5$ of about 1\%, markedly below our method and EVIMO. Overall, our approach improves mIoU over OMS by 61\%. Further videos on the motion segmentation performance of our method on DSEC \cite{gehrig2021dsec} are available in the attached supplementary video.

\subsection{Motion Segmentation on EVIMO \label{sec:evimotest}}

\begin{table}[t]
\footnotesize
\setlength{\tabcolsep}{4pt} 
\caption{Point-based Intersection over Union \% (pIoU) for mask prediction at the current instant on EVIMO dataset. E indicates methods using events while I indicates the method using RGB frames.}
\centering
\begin{tabular}{
m{2.8cm}
>{\centering\arraybackslash}m{0.7cm}
>{\centering\arraybackslash}m{0.7cm}
>{\centering\arraybackslash}m{0.7cm}
>{\centering\arraybackslash}m{0.7cm}
>{\centering\arraybackslash}m{0.7cm}
>{\centering\arraybackslash}m{0.7cm}}
\toprule
\textbf{Algorithm} & \textbf{Input} & Boxes & Floor & Wall & Table & Fast \\
\midrule
Baseline CNN \cite{wang2023evmoseg} & E & 50 & \underline{74} & 60 & 66 & 52 \\
\grayrow
EV-IMO \cite{mitrokhin2019ev} & E  & \underline{70} & 59 & 78 & \underline{79} & \underline{67} \\
EVDodgeNet \cite{sanket2020evdodgenet} & E  & 67 & 61 & 72 & 70 & 60 \\
\grayrow
SpikeMS \cite{parameshwara2021spikems} & E  & 65 & 53 & 63 & 50 & 38 \\
GConv \cite{mitrokhin2020learning} & E  & 60 & 55 & \underline{80} & 51 & 39 \\
\grayrow
EMSGC \cite{zhou2021event} Top 30\% & E  & 24 & 18 & 24 & 55 & 43 \\
EMSGC \cite{zhou2021event} Top 50\% & E  & 14 & 11 & 15 & 36 & 26 \\
\grayrow
Un-EVIMO \cite{wang2023evmoseg} & E  & 45 & 56 & 53 & 50 & 44 \\
FlowSAM \cite{xie2024flowsam} & I  & 12 & 10 & 14 & 11 & 13 \\
\midrule
\grayrow
Ours & E & \textbf{72} & \textbf{86} & \textbf{80} & \textbf{80} & \textbf{67} \\
\bottomrule
\end{tabular}
\label{tab:current_on_evimo}
\end{table}

We further assess our method by comparing it against state-of-the-art approaches on the EVIMO dataset, as summarized in Tab.~\ref{tab:current_on_evimo}. Evaluation follows the protocol established by Un-EVIMO \cite{wang2023evmoseg}, using point-based Intersection over Union (pIoU \%) the same metric adopted in the original EVIMO benchmark \cite{mitrokhin2019ev}.

We compare different baselines including supervised CNN-based methods like EVIMO \cite{mitrokhin2018event}, EVDodgeNet \cite{sanket2020evdodgenet}, spiking neural network based approaches like SpikeMS \cite{parameshwara2021spikems}, graph-based approaches like GConv \cite{mitrokhin2020learning}, unsupervised optimization-based EMSGC \cite{zhou2021event},  Un-EVIMO \cite{wang2023evmoseg} and frame-based approaches like FlowSAM \cite{xie2024flowsam}.



Tab.~\ref{tab:current_on_evimo} demonstrates that our approach outperforms unsupervised approaches by 55\% and spiking-based approaches by 40\%, as well as the state-of-the-art represented by convolutional methods by 10\%.

Furthermore, frame-based methods such as FlowSAM struggle to disentangle complex motion patterns, resulting in low scores on the EVIMO benchmark. We argue that motion segmentation inherently benefits from sensors that measure motion directly; traditional cameras are ill-equipped for this task, as they only capture absolute scene intensity at discrete intervals. Consequently, the results in Tab.~\ref{tab:current_on_evimo} show that FlowSAM suffers a $5.5 \times$ drop in accuracy compared to our method, which relies entirely on a single event camera.

\subsection{Motion Segmentation on EED \label{sec:EEDtest}}
\begin{table}[t]
    \centering
    \begin{minipage}[t]{0.52\columnwidth}
        \centering
        \scriptsize
        \setlength{\tabcolsep}{2pt}
        \caption{Run time [ms], age, \& GFLOPS}
        \label{tab:run_time}
        \begin{tabular}{ l c c c }
            \toprule
            \textbf{Alg.} & \textbf{Time ($\mu$/$\sigma$)} & \textbf{Age} & \textbf{GFLOPS} \\
            \midrule
            EV-IMO & 8.82/0.54 & -8.8 & 1.80 \\
            \grayrow OMS & 112.7/3.5 & -112.7 & \textbf{0.12} \\
            Ours & \textbf{5.76}/0.64 & \textbf{94.2} & 32.15 \\
            \bottomrule
        \end{tabular}
    \end{minipage}%
    \hfill
    \begin{minipage}[t]{0.46\columnwidth}
        \centering
        \scriptsize
        \setlength{\tabcolsep}{2pt} 
        \caption{IMO pred. (mIoU\%) on EED}
        \label{tab:EED_mIoU}
        \begin{tabular}{ l c c c c c }
            \toprule
            \textbf{Alg.} & \textbf{FD} & \textbf{OC} & \textbf{WIB} & \textbf{LV} & \textbf{MO} \\
            \midrule
            EBMS & 27.2 & 69.7 & 68.4 & 31.0 & 41.6 \\
            \grayrow PMS & 53.9 & \textbf{83.2} & \textbf{93.3} & 48.3 & 51.1 \\
            Ours & \textbf{72.6} & 78.5 & 79.6 & \textbf{63.3} & \textbf{55.4} \\
            \bottomrule
        \end{tabular}
    \end{minipage}
\end{table}
To evaluate the robustness of our approach to varying illumination conditions and severe sensor noise, we additionally benchmark our method on the Event-based Ego-motion and Dynamic-object (EED) dataset \cite{EED_mitrokhin2018event}. Unlike standard indoor scenes, EED features highly challenging sequences characterized by low illumination and complex, unconstrained trajectories of multiple independently moving objects. 

We compare our method against two recent event-based motion segmentation approaches: EBMS \cite{stoffregen2019event} and ProgressiveMotionSeg (PMS) \cite{chen2022progressivemotionseg}. As shown in Tab.~\ref{tab:EED_mIoU}, performance is measured using the mean Intersection over Union (mIoU \%) across five distinct sequence categories: Fast Drone (FD), OC, WIB, Light Variation (LV), and Multiple Objects (MO).

Our model demonstrates highly competitive performance, achieving state-of-the-art results in the FD, LV, and MO categories. Specifically, our approach outperforms the recent PMS baseline by a significant margin in low-light conditions (LV: 15.0\%) and scenes with multiple dynamic agents (MO: 4.3\%). This underscores the efficacy of our anticipatory suppression mechanism, which successfully isolates IMOs even when standard brightness constancy assumptions fail or event generation is sparse due to darkness. 

While PMS achieves higher accuracy in the OC and WIB sequences, our method maintains strong, consistent performance across the board (outperforming EBMS in all categories). Ultimately, these results prove that our framework generalizes well beyond standard DSEC driving scenarios and EVIMO tabletop scenes, remaining robust in noisy, real-world dynamic environments.

\subsection{Run time, Prediction Age and GFLOPS Evaluations \label{timecomparison}}

We evaluate the time complexity of our approach by comparing wall-clock inference time, prediction "age," and GFLOPS against baselines on a GeForce RTX 2080 Ti and 1 core of a consumer CPU. As detailed in Tab.~\ref{tab:run_time}, our method achieves the fastest processing time of $5.8$ms ($\approx 173$\,Hz), which is $\approx 53\%$ faster than EV-IMO \cite{mitrokhin2019ev} and two orders of magnitude faster than the open-source implementation of OMS \cite{clerico2408retina}.

Crucially, the baseline methods predict masks that lag behind the target timestep due to their respective runtime (negative prediction age).
Instead, our method sets forecasts by 100\,ms, reaching a prediction age of 94.2\,ms. This successfully compensates for the computation time of $5.8$ms, delivering zero-latency masks effectively. Interestingly, although our model incurs higher theoretical complexity (32 GFLOPS) compared to EV-IMO (1.8) and OMS (0.12), it yields the lowest latency. This disparity highlights the efficiency of GPU-optimized architectures over CPU-bound, hand-tuned algorithms like OMS, where low flop counts do not translate to real-time performance.
Details of the experiment are available in the Appendix (Sec. \ref{sec:appendix_time_comparison}).

\begin{figure}[t]
    \centering
    \includegraphics[width=\columnwidth]{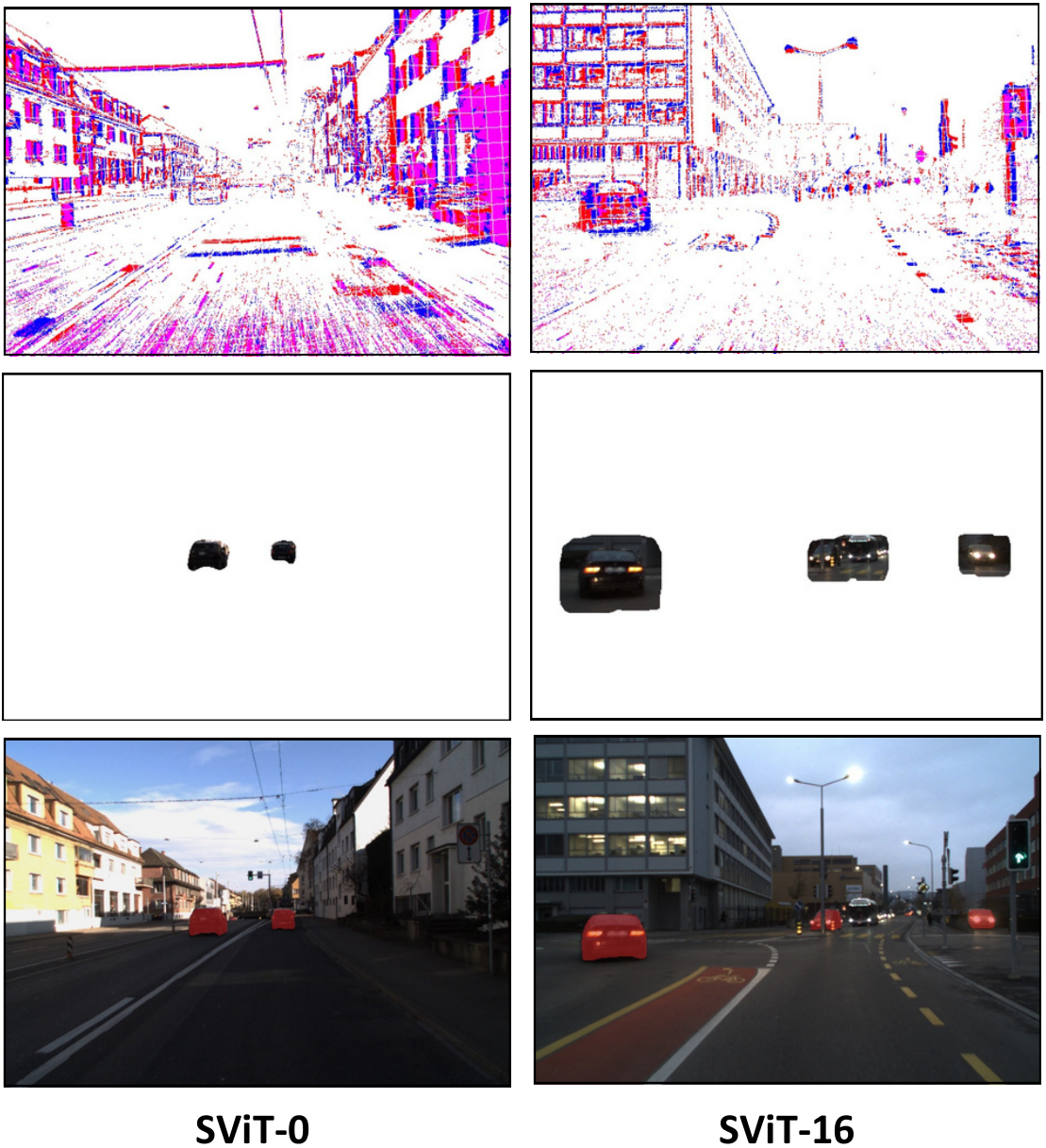}
    \captionof{figure}{
        Dynamic object masks for SViT token pruning. Left to right: SViT-0 and SViT-16 show increasing mask dilation. Larger dilation factors cover more area, including more background tokens and increasing system latency. This mask expansion introduces redundancy that significantly reduces inference frequency while capturing the same dynamic content of the scene.}
    \label{fig:svit_showcase}
\end{figure}

\subsection{Ablations \label{ablations}}

\begin{table}[t]
\footnotesize
\setlength{\tabcolsep}{4pt} 
\caption{Mean Intersection Over Union \% (mIoU) Comparison of Core Components of Our Framework on EVIMO dataset}
\centering
\begin{tabular}{m{0.7cm}>
{\centering\arraybackslash}m{0.7cm}>
{\centering\arraybackslash}m{0.7cm}>
{\centering\arraybackslash}m{0.7cm}>
{\centering\arraybackslash}m{0.7cm}>
{\centering\arraybackslash}m{0.7cm}>
{\centering\arraybackslash}m{0.7cm}}
    \textbf{FW} & \textbf{ATC} & Boxes & Floor & Wall & Table & Fast \\
    \midrule
    
    $\times$ & $\times$             & 72.25 & 73.08 & 68.62 & 78.71 & 65.06\\
    \grayrow
    Linear & $\times$               & 73.88 & 75.41 & 71.01 & 78.02 & 61.77\\
    
    $\checkmark$ & $\times$         & 74.03 & 78.02 & 75.14 & 77.70 & 67.49\\
    \grayrow
    $\checkmark$ & $\checkmark$     & \textbf{76.24} & \textbf{80.10} & \textbf{79.63} & \textbf{79.71} & \textbf{68.07}\\
    \vspace{0pt}
\end{tabular}
\label{tab:algorithm_comparison}
\end{table}

Unlike driving scenarios, indoor EVIMO scenes feature highly non-linear object motion, causing segmentation masks to lag behind the true object position within milliseconds. We quantify the necessity of our non-linear forecasting by comparing our full model (ATC + Flow Warping) against three baselines: (1) no forecasting, (2) linear motion extrapolation, and (3) basic temporal summation without ATC.

As shown in Tab.~\ref{tab:algorithm_comparison}, linear extrapolation improves upon the static baseline but fails in sequences with chaotic motion, such as Floor and Fast. Our learned flow warping significantly outperforms these simpler alternatives, confirming that forecasting curved pixel paths is essential for robust anticipatory segmentation.

\begin{table}[t]
\footnotesize
\setlength{\tabcolsep}{4pt} 
\caption{Mean Intersection Over Union \% (mIoU) for Our Model at Different Scales On EVIMO dataset}
\centering
\begin{tabular}{m{0.7cm}>
{\centering\arraybackslash}m{0.7cm}>
{\centering\arraybackslash}m{0.7cm}>
{\centering\arraybackslash}m{0.7cm}>
{\centering\arraybackslash}m{0.7cm}>
{\centering\arraybackslash}m{0.7cm}}
    \textbf{Model} & Boxes & Floor & Wall & Table & Fast \\
    \midrule
    
    XS     & 47.12 & 47.82 & 46.26 & 47.57 & 50.74 \\
    \grayrow
    S      & 68.72 & 80.07 & 75.96 & 73.93 & 58.52 \\
    
    M      & 72.68 & 80.00 & 75.04 & 79.02 & 64.59 \\
    \grayrow
    L      & \textbf{76.24} & \textbf{80.10} & \textbf{79.63} & \textbf{79.71} & \textbf{68.07}
\end{tabular}
\label{tab:ablation_model_size}
\end{table}

We further analyze the impact of model capacity in Tab.~\ref{tab:ablation_model_size} by scaling the architecture from 25\% (XS) to 100\% (L) of the total parameters. Results demonstrate that increasing model size yields consistent mIoU gains, particularly in highly dynamic scenes. Although the improvement \new{from Small to Large and} from Medium to Large is marginal, the Large model achieves peak accuracy while maintaining the real-time performance established in Sec.~\ref{timecomparison}.

Further \new{experiments} on the system sensitivity to the event rate are available in the appendix (Sec. \ref{sec:appendix_sensitivity_rate}).

\subsection{Accuracy vs. Prediction Horizon Ablation\label{sec:pred_horizon_ablation}}
    We study how segmentation accuracy varies with the prediction horizon by training the small model "S" while randomizing the prediction time during training. Given a set of horizons, we uniformly sample a $\Delta t_p \in [0,100]ms$ and supervise the model using events from a fixed window of length $\Delta t$ such that the supervision time is $\Delta t+\Delta t_p$ and aligns with the ground-truth timestamp. This avoids any ground-truth interpolation.
    
\begin{figure}[t]
    \centering
    \includegraphics[width=\columnwidth]{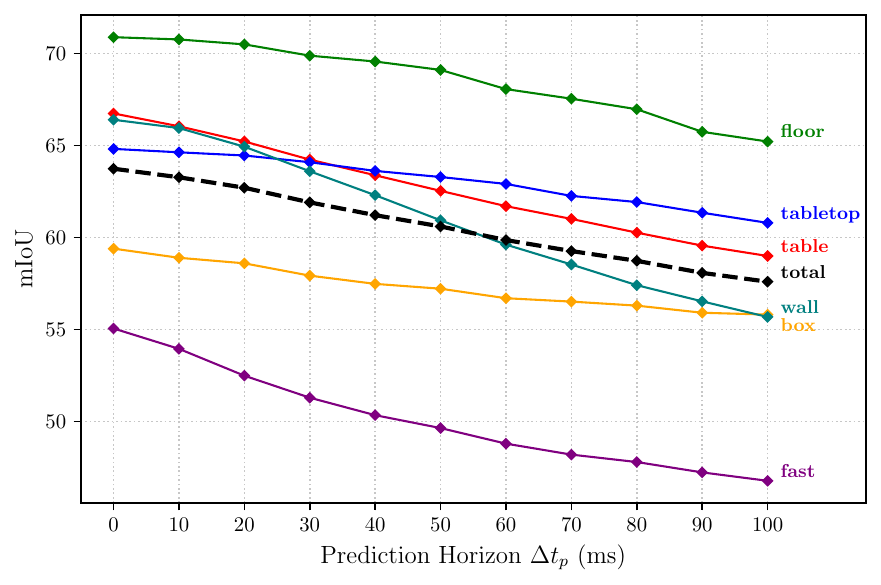}
        \captionof{figure}{Mean IoU (mIoU) versus prediction horizon $\Delta t_p$ (0--100\,ms) on the EVIMO dataset for six sequences (box, fast, floor, table, tabletop, wall) and the macro-average total. Performance declines with increasing horizon, indicating that longer look-ahead yields less accurate segmentation. Faster scenes such as wall and fast degrade more rapidly than slower ones like box and tabletop, due to stronger non-smooth motion and occlusions that increase the effective prediction age.
        }
    \label{fig:time_ablation}
\end{figure}

    We report mean IoU (mIoU) for each fixed horizon on six EVIMO sequences (box, fast, floor, table, tabletop, wall) and the performance averaged over all scenes (total). Results in Fig.~\ref{fig:time_ablation} show a consistent, monotonic drop in mIoU as $\Delta t_p$ increases, reflecting the growing difficulty of extrapolating fine object boundaries further into the future. The decay rate is scene-dependent: rapid, non-smooth sequences like fast and wall degrade more steeply, due to occlusions and larger relative pose changes, whereas slower scenes like box and tabletop degrade more gradually. Thus, while horizon randomization improves robustness across time offsets, performance remains, as expected, limited by the prediction horizon; in latency-sensitive applications, shorter $\Delta t_p$ should be preferred to mitigate such errors.

\begin{table}[t]
\footnotesize
\setlength{\tabcolsep}{4pt} 
\caption{ATE [m] (Absolute Trajectory Error) Comparison of RAMP-VO on EVIMO Dataset}
\centering
\begin{tabular}{
m{2.0cm}
>{\centering\arraybackslash}m{0.6cm}
>{\centering\arraybackslash}m{0.6cm}
>{\centering\arraybackslash}m{0.6cm}
>{\centering\arraybackslash}m{0.6cm}
>{\centering\arraybackslash}m{0.6cm}
>{\centering\arraybackslash}m{0.6cm}}
\textbf{Algorithm} & Boxes & Floor & Wall & Table & Fast & Tabletop \\
\midrule

RAMP-VO \cite{Pellerito_2024_IROS} & 0.33 & 0.17 & 0.34 & \underline{0.15} & 0.25 & 0.14 \\
\grayrow
RAMP-VO$^*$ \cite{Pellerito_2024_IROS} & \underline{0.29} & \underline{0.15} & \underline{0.33} & \underline{0.15} & \underline{0.22} & \underline{0.12} \\

RAMP-VO$^\dagger$ \cite{Pellerito_2024_IROS} & 0.28 & 0.15 & 0.30 & 0.15 & 0.22 & 0.11 \\
\vspace{0pt}
\end{tabular}
\label{tab:application_vo}
\end{table}

\subsection{Application on Visual Odometry \label{visualodometry}}

We evaluate the impact of anticipatory event suppression on state estimation by integrating our method into an open-source event- and frame based VO pipeline RAMP-VO \cite{Pellerito_2024_IROS}. Experiments on EVIMO \cite{mitrokhin2019ev} compare three configurations: standard RAMP-VO, RAMP-VO$^*$ (using our suppressor), and RAMP-VO$^\dagger$ (using ground-truth masks as an upper bound). Our method enhances robustness in dynamic scenes by lowering the mean Absolute Trajectory Error (ATE) to 0.21\,m ($-8.7\%$), effectively bridging the gap toward the maximum achievable gain of $-13\%$ demonstrated by RAMP-VO$^\dagger$. The improvement is most pronounced on dynamic sequences (Boxes, Wall, Fast) where IMO features typically corrupt pose estimation, while performance remains stable on static scenes, confirming that filtering does not degrade valid ego-motion data.

\begin{figure}[t]
    \centering
    \includegraphics[width=\columnwidth]{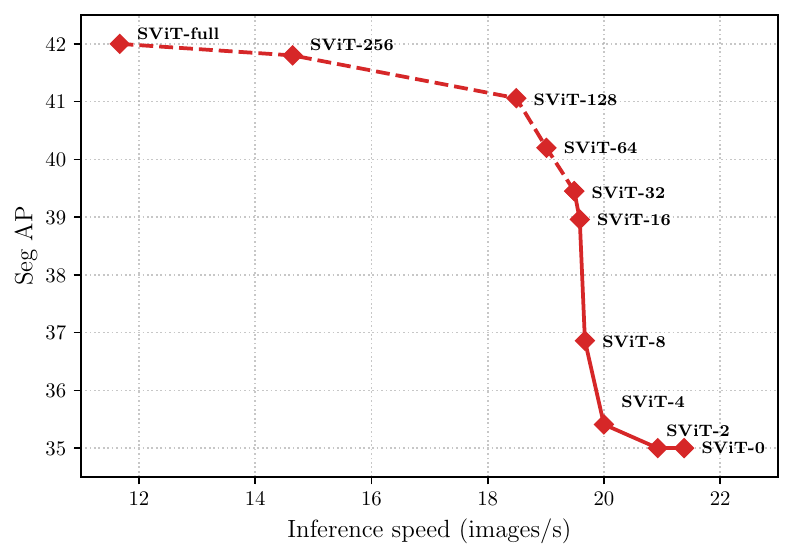} 
    \captionof{figure}{%
    Trade-off between segmentation accuracy (Seg AP) and inference frequency for different pruning ratios of the SViT model. As the pruning ratio increases (e.g., from SViT-full to SViT-0), the inference frequency improves, but the segmentation performance degrades.}
    \label{fig:application_token_pruning}
\end{figure}

\subsection{Application on Token pruning for Transformers Efficiency \label{tokenpruning}}
Vision Transformers process an image as a flat sequence of patch tokens, which makes the inference cost grow quadratically with the number of tokens. Earlier work focused on pruning tokens uniformly at random or using simplified gating mechanisms, as in \cite{liu2024revisiting}. 
We propose using event-based anticipatory masks to selectively prune static-scene tokens, focusing computation on dynamic agents (e.g., pedestrians, vehicles). 
By substituting the gating mechanism in SViT \cite{liu2024revisiting} with our dynamic masks on the DSEC dataset, we achieve a superior efficiency-accuracy balance without incurring extra time costs, given that the masks are forecasted in advance.
We simulate the impact of increasingly coarser IMO segmentations by dilating the original mask estimated by our model (SVIT-0). As shown in Fig.~\ref{fig:svit_showcase}, coarser masks incur higher token utilization.
As shown in Fig.~\ref{fig:application_token_pruning}, using our original mask (SViT-0) increases inference speed by approximately 10~FPS compared to the full-token baseline (SViT-full) \cite{liu2024revisiting}, with a negligible drop of less than 7 points in Segmentation Average Precision (SegAP).
Further videos on token pruning performance are available in the attached supplementary video.

\section{Limitations}
\label{sec:limitations}

Despite its effectiveness, our approach has several limitations.
While the method generalizes well across varied camera resolutions and low-light conditions, our anticipatory suppression relies on accurate flow forecasting and mask warping. Although it successfully handles small ego-motion (1-2 px/frame), under larger ego-motion, abrupt non-linear motion, fast occlusions, or depth-layer changes, the warped mask can misalign and blur boundaries, reducing IoU.
Moreover, the “anticipation” effect depends on runtime being small relative to the chosen forecast horizon; therefore, if compute budgets shrink or horizons grow, the effective prediction age advantage narrows.
Performance also degrades as the prediction horizon increases, with steeper drops in highly dynamic scenes, which limits long motion forecasting settings.
Critically, accuracy is sensitive to event density. Our method is most robust when processing between $2 \times 10^3$ and $10^6$ events; outside this regime, the method struggles. IMOs that generate very few events (e.g., distant cars or pedestrians) do not provide enough context to be distinguished from the background, while excessively dense windows integrating over long spans can blur object boundaries.
Finally, while our approach natively predicts an arbitrary number of IMOs without periodic re-initialization, our evaluation currently emphasizes EVIMO (indoor) and DSEC (driving). Broader validation on egocentric manipulation, legged platforms, and aerial viewpoints remains open. Future works should include uncertainty-aware warping and evaluation across a wider range of robotic platforms and environments.

\section{Conclusion}
In this work, we introduced a novel framework for real-time Anticipatory Motion Suppression, directly addressing the critical challenge of filtering irrelevant motion activity from event streams to enable more efficient downstream perception. Our key innovation is a unified, end-to-end architecture that breaks the traditional trade-off between accuracy and latency by jointly learning to segment independently moving objects and predict their future motion. 

We have validated our approach on the challenging EVIMO benchmark, where it establishes a new state of the art, outperforming prior methods by a significant margin of 67\% while operating 60 Hz faster. Our model's efficiency running at 173 Hz on a consumer GPU demonstrate its suitability for real-world, resource constrained applications. Moreover, we show the profound practical impact of our method. By supplying a clean stream of ego-motion only events, we improved the accuracy of a state-of-the-art visual odometry system by 13\%. Furthermore, by using the IMOs masks to guide token pruning, we accelerated Vision Transformer's inference speed by 10 FPS with only a negligible drop in accuracy. Our work represents a significant step towards making event-based perception more robust, efficient, and practical for latency-critical robotic systems like autonomous vehicles and AR/VR devices.

\clearpage

\bibliographystyle{IEEEtran}  
\bibliography{references.bib}

@article{gehrig2021dsec,
  title={Dsec: A stereo event camera dataset for driving scenarios},
  author={Gehrig, Mathias and Aarents, Willem and Gehrig, Daniel and Scaramuzza, Davide},
  journal={IEEE Robotics and Automation Letters},
  volume={6},
  number={3},
  pages={4947--4954},
  year={2021},
  publisher={IEEE}
}

@inproceedings{wang2023evmoseg,
  title={Un-EVIMO: Unsupervised event-based independent motion segmentation},
  author={Wang, Ziyun and Guo, Jinyuan and Daniilidis, Kostas},
  booktitle={European Conference on Computer Vision},
  pages={228--245},
  year={2024},
  organization={Springer}
}

@article{zhang2023multi,
  title={A multi-scale recurrent framework for motion segmentation with event camera},
  author={Zhang, Shaobo and Sun, Lei and Wang, Kaiwei},
  journal={IEEE Access},
  volume={11},
  pages={80105--80114},
  year={2023},
  publisher={IEEE}
}

@inproceedings{georgoulis2024out,
  title={Out of the Room: Generalizing Event-Based Dynamic Motion Segmentation for Complex Scenes},
  author={Georgoulis, Stamatios and Ren, Weining and Bochicchio, Alfredo and Eckert, Daniel and Li, Yuanyou and Gawel, Abel},
  booktitle={2024 International Conference on 3D Vision (3DV)},
  pages={442--452},
  year={2024},
  organization={IEEE}
}

@inproceedings{stoffregen2019event,
  title={Event-based motion segmentation by motion compensation},
  author={Stoffregen, Timo and Gallego, Guillermo and Drummond, Tom and Kleeman, Lindsay and Scaramuzza, Davide},
  booktitle={Proceedings of the IEEE/CVF International Conference on Computer Vision},
  pages={7244--7253},
  year={2019}
}

@inproceedings{mitrokhin2018event,
  title={Event-based moving object detection and tracking},
  author={Mitrokhin, Anton and Ferm{\"u}ller, Cornelia and Parameshwara, Chethan and Aloimonos, Yiannis},
  booktitle={2018 IEEE/RSJ International Conference on Intelligent Robots and Systems (IROS)},
  pages={1--9},
  year={2018},
  organization={IEEE}
}

@inproceedings{mitrokhin2019ev,
  title={EV-IMO: Motion segmentation dataset and learning pipeline for event cameras},
  author={Mitrokhin, Anton and Ye, Chengxi and Ferm{\"u}ller, Cornelia and Aloimonos, Yiannis and Delbruck, Tobi},
  booktitle={2019 IEEE/RSJ International Conference on Intelligent Robots and Systems (IROS)},
  pages={6105--6112},
  year={2019},
  organization={IEEE}
}

@inproceedings{parameshwara2021spikems,
  title={SpikeMS: Deep spiking neural network for motion segmentation},
  author={Parameshwara, Chethan M and Li, Simin and Ferm{\"u}ller, Cornelia and Sanket, Nitin J and Evanusa, Matthew S and Aloimonos, Yiannis},
  booktitle={2021 IEEE/RSJ International Conference on Intelligent Robots and Systems (IROS)},
  pages={3414--3420},
  year={2021},
  organization={IEEE}
}

@article{clerico2408retina,
  title={Retina-Inspired Object Motion Segmentation for Event-Cameras},
  author={Clerico, Victoria and Snyder, Shay and Lohia, Arya and Abdullah-Al Kaiser, Md and Schwartz, Gregory and Jaiswal, Akhilesh and Parsa, Maryam},
  journal={arXiv preprint arXiv:2408.09454}
}

@inproceedings{mitrokhin2020learning,
  title={Learning visual motion segmentation using event surfaces},
  author={Mitrokhin, Anton and Hua, Zhiyuan and Fermuller, Cornelia and Aloimonos, Yiannis},
  booktitle={Proceedings of the IEEE/CVF Conference on Computer Vision and Pattern Recognition},
  pages={14414--14423},
  year={2020}
}

@article{zhou2021event,
  title={Event-based motion segmentation with spatio-temporal graph cuts},
  author={Zhou, Yi and Gallego, Guillermo and Lu, Xiuyuan and Liu, Siqi and Shen, Shaojie},
  journal={IEEE transactions on neural networks and learning systems},
  volume={34},
  number={8},
  pages={4868--4880},
  year={2021},
  publisher={IEEE}
}

@inproceedings{sanket2020evdodgenet,
  title={Evdodgenet: Deep dynamic obstacle dodging with event cameras},
  author={Sanket, Nitin J and Parameshwara, Chethan M and Singh, Chahat Deep and Kuruttukulam, Ashwin V and Ferm{\"u}ller, Cornelia and Scaramuzza, Davide and Aloimonos, Yiannis},
  booktitle={2020 IEEE International Conference on Robotics and Automation (ICRA)},
  pages={10651--10657},
  year={2020},
  organization={IEEE}
}

@inproceedings{parameshwara20210,
  title={0-mms: Zero-shot multi-motion segmentation with a monocular event camera},
  author={Parameshwara, Chethan M and Sanket, Nitin J and Singh, Chahat Deep and Ferm{\"u}ller, Cornelia and Aloimonos, Yiannis},
  booktitle={2021 IEEE International Conference on Robotics and Automation (ICRA)},
  pages={9594--9600},
  year={2021},
  organization={IEEE}
}

@article{falanga2020dynamic,
  title={Dynamic obstacle avoidance for quadrotors with event cameras},
  author={Falanga, Davide and Kleber, Kevin and Scaramuzza, Davide},
  journal={Science Robotics},
  volume={5},
  number={40},
  pages={eaaz9712},
  year={2020},
  publisher={American Association for the Advancement of Science}
}

@InProceedings{Pellerito_2024_IROS,
    author    = {Pellerito, Roberto and Cannici, Marco and Gehrig, Daniel and Belhadj, Joris and 
                Dubois-Matra, Olivier and Casasco, Massimo and Scaramuzza, Davide},
    title     = {Deep Visual Odometry with Events and Frames},
    booktitle = {IEEE/RSJ International Conference on Intelligent Robots (IROS)},
    month     = {June},
    year      = {2024}
}

@inproceedings{stoffregen2022event,
  title={Event-based kilohertz eye tracking using coded differential lighting},
  author={Stoffregen, Timo and Daraei, Hossein and Robinson, Clare and Fix, Alexander},
  booktitle={Proceedings of the IEEE/CVF Winter Conference on Applications of Computer Vision},
  pages={2515--2523},
  year={2022}
}

@article{gehrig2024low,
  title={Low-latency automotive vision with event cameras},
  author={Gehrig, Daniel and Scaramuzza, Davide},
  journal={Nature},
  volume={629},
  number={8014},
  pages={1034--1040},
  year={2024},
  publisher={Nature Publishing Group UK London}
}

@article{mostafavi2021learning,
  title={Learning to reconstruct hdr images from events, with applications to depth and flow prediction},
  author={Mostafavi, Mohammad and Wang, Lin and Yoon, Kuk-Jin},
  journal={International Journal of Computer Vision},
  volume={129},
  number={4},
  pages={900--920},
  year={2021},
  publisher={Springer}
}

@InProceedings{Paredes-Valles_2023_ICCV,
    author    = {Paredes-Vall\'es, Federico and Scheper, Kirk Y. W. and De Wagter, Christophe and de Croon, Guido C. H. E.},
    title     = {Taming Contrast Maximization for Learning Sequential, Low-latency, Event-based Optical Flow},
    booktitle = {Proceedings of the IEEE/CVF International Conference on Computer Vision (ICCV)},
    month     = {October},
    year      = {2023},
    pages     = {9695-9705}
}

@article{charbonnier1997deterministic,
  title={Deterministic edge-preserving regularization in computed imaging},
  author={Charbonnier, Pierre and Blanc-F{\'e}raud, Laure and Aubert, Gilles and Barlaud, Michel},
  journal={IEEE Transactions on image processing},
  volume={6},
  number={2},
  pages={298--311},
  year={1997},
  publisher={IEEE}
}

@InProceedings{Gallego_2019_CVPR,
author = {Gallego, Guillermo and Gehrig, Mathias and Scaramuzza, Davide},
title = {Focus Is All You Need: Loss Functions for Event-Based Vision},
booktitle = {Proceedings of the IEEE/CVF Conference on Computer Vision and Pattern Recognition (CVPR)},
month = {June},
year = {2019}
}

@InProceedings{Gallego_2018_CVPR,
author = {Gallego, Guillermo and Rebecq, Henri and Scaramuzza, Davide},
title = {A Unifying Contrast Maximization Framework for Event Cameras, With Applications to Motion, Depth, and Optical Flow Estimation},
booktitle = {Proceedings of the IEEE Conference on Computer Vision and Pattern Recognition (CVPR)},
month = {June},
year = {2018}
}

@inproceedings{liu2024revisiting,
  title={Revisiting token pruning for object detection and instance segmentation},
  author={Liu, Yifei and Gehrig, Mathias and Messikommer, Nico and Cannici, Marco and Scaramuzza, Davide},
  booktitle={Proceedings of the IEEE/CVF Winter Conference on Applications of Computer Vision},
  pages={2658--2668},
  year={2024}
}

@article{zhu2018ev,
  title={EV-FlowNet: Self-supervised optical flow estimation for event-based cameras},
  author={Zhu, Alex Zihao and Yuan, Liangzhe and Chaney, Kenneth and Daniilidis, Kostas},
  journal={arXiv preprint arXiv:1802.06898},
  year={2018}
}

@article{ye2018unsupervised,
  title={Unsupervised learning of dense optical flow, depth and egomotion from sparse event data},
  author={Ye, Chengxi and Mitrokhin, Anton and Ferm{\"u}ller, Cornelia and Yorke, James A and Aloimonos, Yiannis},
  journal={arXiv preprint arXiv:1809.08625},
  year={2018}
}

@inproceedings{gehrig2021raft,
  title={E-raft: Dense optical flow from event cameras},
  author={Gehrig, Mathias and Millh{\"a}usler, Mario and Gehrig, Daniel and Scaramuzza, Davide},
  booktitle={2021 International Conference on 3D Vision (3DV)},
  pages={197--206},
  year={2021},
  organization={IEEE}
}

@inproceedings{paredes2023taming,
  title={Taming contrast maximization for learning sequential, low-latency, event-based optical flow},
  author={Paredes-Vall{\'e}s, Federico and Scheper, Kirk YW and De Wagter, Christophe and De Croon, Guido CHE},
  booktitle={Proceedings of the IEEE/CVF International Conference on Computer Vision},
  pages={9695--9705},
  year={2023}
}

@inproceedings{shiba2022secrets,
  title={Secrets of event-based optical flow},
  author={Shiba, Shintaro and Aoki, Yoshimitsu and Gallego, Guillermo},
  booktitle={European Conference on Computer Vision},
  pages={628--645},
  year={2022},
  organization={Springer}
}

@article{vaswani2017attention,
  title={Attention is all you need},
  author={Vaswani, Ashish and Shazeer, Noam and Parmar, Niki and Uszkoreit, Jakob and Jones, Llion and Gomez, Aidan N and Kaiser, {\L}ukasz and Polosukhin, Illia},
  journal={Advances in neural information processing systems},
  volume={30},
  year={2017}
}

@article{li2019dice,
  title={Dice loss for data-imbalanced NLP tasks},
  author={Li, Xiaoya and Sun, Xiaofei and Meng, Yuxian and Liang, Junjun and Wu, Fei and Li, Jiwei},
  journal={arXiv preprint arXiv:1911.02855},
  year={2019}
}

@article{d2025wandering,
  title={Wandering around: A bioinspired approach to visual attention through object motion sensitivity},
  author={D’Angelo, Giulia and Clerico, Victoria and Bartolozzi, Chiara and Hoffmann, Matej and Furlong, P Michael and Hadjiivanov, Alexander},
  journal={Neuromorphic Computing and Engineering},
  volume={5},
  number={2},
  pages={024019},
  year={2025},
  publisher={IOP Publishing}
}

@article{ballas2015delving,
  title={Delving deeper into convolutional networks for learning video representations},
  author={Ballas, Nicolas and Yao, Li and Pal, Chris and Courville, Aaron},
  journal={arXiv preprint arXiv:1511.06432},
  year={2015}
}

@InProceedings{xie2024flowsam,
  title={Moving Object Segmentation: All You Need Is SAM (and Flow)},
  author={Junyu Xie and Charig Yang and Weidi Xie and Andrew Zisserman},
  booktitle={ACCV},
  year={2024}
}

@inproceedings{EED_mitrokhin2018event,
  title={Event-based moving object detection and tracking},
  author={Mitrokhin, Anton and Ferm{\"u}ller, Cornelia and Parameshwara, Chethan and Aloimonos, Yiannis},
  booktitle={2018 IEEE/RSJ International Conference on Intelligent Robots and Systems (IROS)},
  pages={1--9},
  year={2018},
  organization={IEEE}
}

@inproceedings{chen2022progressivemotionseg,
  title={Progressivemotionseg: Mutually reinforced framework for event-based motion segmentation},
  author={Chen, Jinze and Wang, Yang and Cao, Yang and Wu, Feng and Zha, Zheng-Jun},
  booktitle={Proceedings of the AAAI Conference on Artificial Intelligence},
  volume={36},
  number={1},
  pages={303--311},
  year={2022}
}

\clearpage
\appendices 

\twocolumn[
  \begin{center}
    \Huge Appendix for:\\
    Motion-aware Event Suppression for Event Cameras \\
    \vspace{0.5cm}
    \large Roberto Pellerito, Nico Messikommer, Giovanni Cioffi, Marco Cannici, Davide Scaramuzza\\ Robotics and Perception Group, University of Zurich, Switzerland \\
    \vspace{0.5cm}
    \normalsize
  \end{center}
]


\section{Method Details}
\subsection{Attention-based Time Conditioning (ATC)\label{sec:appendix_timeconditioning}}
    To enable the forecasting of optical flow at a future time step $t + \Delta t_p$ we introduce a temporal attention module. This module's purpose is to dynamically modulate the current spatial feature embedding, $\mathbf{E} \in \mathbb{R}^{B \times C \times H \times W}$, based on the target time delta. It achieves this using a multi-head cross-attention mechanism, depicted in Fig. ~\ref{fig:atc} where the Query is derived from the time information, while the Key and Value are derived from the spatial features. This process is composed of three main steps: 
    
\paragraph{Preparation of Attention Inputs}
    First, the temporal and spatial inputs must be transformed into the sequence format required by the attention mechanism, which is `[Batch, Sequence-Length, Embedding-Dim]`.
    
    \begin{itemize}
        \item Query Tensor ($\mathbf{Q}$): The scalar time delta, $\Delta t$, is first encoded into a $C$-dimensional vector using the sinusoidal positional encoding function, yielding $\mathbf{t}_{\text{enc}} = \text{PE}(\Delta t) \in \mathbb{R}^{B \times C}$. More details on the positional encoding can be found in ~\ref{sec:appendix_positionalencoding}. This single vector is then broadcast across all $H \times W$ spatial positions to form the Query tensor $\mathbf{Q} \in \mathbb{R}^{B \times (H \cdot W) \times C}$. This implies that every spatial location is being queried with the same temporal information.
    
        \item Key and Value Tensors ($\mathbf{K}, \mathbf{V}$): The spatial feature map $\mathbf{E}$ is unrolled into a sequence of per-pixel feature vectors. This is achieved by flattening the spatial dimensions, transforming the tensor from shape $(B, C, H, W)$ to $(B, H \cdot W, C)$. This resulting sequence serves as both the Key, $\mathbf{K}$, and the Value, $\mathbf{V}$.
        \begin{align}
            \mathbf{K}, \mathbf{V} \in \mathbb{R}^{B \times (H \cdot W) \times C}
        \end{align}
    \end{itemize}

\paragraph{Multi-Head Cross-Attention}
    The operation computes a weighted sum of the Value vectors, where the weights are determined by the compatibility between the time-derived Query and the space-derived Key. The output is a new sequence of feature vectors, $\mathbf{E}' \in \mathbb{R}^{B \times (H \cdot W) \times C}$, which represents a version of the original spatial features modulated by the temporal query.

\paragraph{Output Reshaping}
    The output sequence $\mathbf{E}'$ is reshaped back into its original spatial tensor format. This "un-flattening" operation transforms the tensor from $(B, H \cdot W, C)$ back to $(B, C, H, W)$. The resulting tensor, $\mathbf{E}_{t+\Delta t}$, is the time-conditioned embedding, which is to be processed by the optical flow decoder.

\subsection{Temporal Positional Encoding\label{sec:appendix_positionalencoding}}
    To provide the model with a continuous notion of time, we inject a temporal encoding for each time scalar $\Delta t$. We adapt the sinusoidal positional encoding scheme from the original Transformer architecture \cite{vaswani2017attention}. However, instead of encoding discrete integer positions, our method encodes the continuous time value $\Delta t$ into a high-dimensional vector. 
    
    Given an embedding dimension $d$, the temporal encoding $\text{PE}(\Delta t) \in \mathbb{R}^d$ is computed as follows:
    \begin{equation}
    \label{eq:temporal_encoding}
    \text{PE}(\Delta t, i) = 
    \begin{cases} 
          \sin(\omega_k \cdot \Delta t) & \text{if } i = 2k \\
          \cos(\omega_k \cdot \Delta t) & \text{if } i = 2k+1 \\
    \end{cases}
    \end{equation}
    where $i$ is the dimension index, and the frequencies $\omega_k$ are defined as:
    \begin{equation}
    \omega_k = \frac{1}{10000^{2k/d}} \quad \text{for } k \in [0, \dots, d/2-1].
    \end{equation}
    This function maps the single time value $\Delta t$ into a $d$-dimensional vector using a bank of sinusoids with geometrically spaced frequencies. This technique provides a rich, continuous, and unambiguous representation of time, allowing the model to easily reason about and generalize to different temporal intervals, a capability not afforded by feeding in the raw scalar value directly.

\subsection{Training Losses\label{sec:appendix_losses}}
    Our network is trained end-to-end using a hybrid, multi-task loss function, which is a weighted sum of a supervised component, $\mathcal{L}_{\text{sup}}$, and an unsupervised component, $\mathcal{L}_{\text{unsup}}$. This hybrid approach leverages available ground truth data while also benefiting from a dense, self-supervised signal derived directly from the raw event data. The total loss is formulated as:
    \begin{equation}
    \label{eq:total_loss}
    \mathcal{L}_{\text{total}} = w_{\text{sup}}\mathcal{L}_{\text{sup}} + w_{\text{unsup}}\mathcal{L}_{\text{unsup}}
    \end{equation}
    where $w_{\text{sup}}$ and $w_{\text{unsup}}$ are scalar weights balancing the two components.

\subsubsection{Supervised Loss Component ($\mathcal{L}_{\text{sup}}$)}
    The supervised loss term is a combination of losses for the dynamic object segmentation and optical flow prediction tasks, applied wherever ground truth is available.

\paragraph{Segmentation Loss ($\mathcal{L}_{\text{mask}}$)}
    For the segmentation task, we use a combined Binary Cross-Entropy (BCE) and Dice loss \cite{li2019dice}, which is effective for handling class imbalance in segmentation problems. The loss is computed for the predicted masks at the current time, $M_t$, and optionally for the future predicted masks, $M_{t+\Delta t}$:
    
    \begin{multline}\label{eq:mask-loss}
    \mathcal{L}_{\text{mask}} =
    \mathcal{L}_{\text{BCE-Dice}}(M_t, M'_t) + \\
    + \lambda_{\text{future\_mask}}
      \mathcal{L}_{\text{BCE-Dice}}(M_{t+\Delta t}, M'_{t+\Delta t})
    \end{multline}
    
    where $M'$ denotes the ground truth mask and $\lambda_{\text{future\_mask}}$ is a weighting factor for the future prediction. The $\mathcal{L}_{\text{BCE-Dice}}$ is a weighted sum of the two individual losses.

\paragraph{Supervised Flow Loss ($\mathcal{L}_{\text{flow\_sup}}$)}
    When ground truth optical flow, $\psi'_{t+\Delta t_p}$, is available, we apply a supervised loss to the predicted flow field $\psi_{t+\Delta t}$. 
    Furthermore, to introduce an auxiliary task, we also predict the current flow field  $\psi_t$ with an additional flow decoder, which is supervised with the ground truth flows $\psi'_t$. 
    For each flow, the corresponding loss consists of two parts: an L1-norm data term computed only on valid pixels, and a spatial smoothness term to encourage locally consistent flow fields. The smoothness is enforced using the Charbonnier loss \cite{charbonnier1997deterministic}, a robust L1-like penalty, on the first-order spatial gradients of the flow field:
    \begin{equation}
    \mathcal{L}_{\text{smooth}}(\psi) = \sum_{p \in \text{pixels}} \sqrt{\|\nabla_x \psi(p)\|^2 + \epsilon^2} + \sqrt{\|\nabla_y \psi(p)\|^2 + \epsilon^2}
    \end{equation}
    The total supervised flow loss is:
    \begin{equation}
    \mathcal{L}_{\text{flow\_sup}} = \lambda_{\text{flow}}\|\psi - \psi'\|_1 + \lambda_{\text{smooth}}\mathcal{L}_{\text{smooth}}(\psi)
    \end{equation}
    where $\lambda_{\text{flow}}$ and $\lambda_{\text{smooth}}$ are hyperparameters.

\subsubsection{Unsupervised Flow Loss ($\mathcal{L}_{\text{unsup}}$)}
    For the optical flow forecasting task, we incorporate a powerful, unsupervised loss based on the principle of contrast maximization ~\cite{Gallego_2019_CVPR} and inspired by \cite{Paredes-Valles_2023_ICCV}. The core idea is that a correct optical flow field, when used to warp events over a time interval, will "deblur" the event stream by mapping moving points back to their origin. 
    In practice, the events are transported to a given reference time $t_{ref}$ applying iteratively the following equation for every event.
    
    \begin{equation}
    \bm{x'_i} = \bm{x_i} + (t_{ref} - t_i)\bm{u(x_i)}
    \end{equation}
    
    Here, $\bm{x'_i}$ represent the $x'_i, y'_i$ coordinates of the transported events and $\bm{x_i}$ the original coordinates. 
    The optimal optical flow $\bm{u(x_i)}$ will warp each event to a location $\bm{x'_i}$, such that the resulting Image of Warped Events (IWE) reaches the maximum contrast, as described in~\cite{Gallego_2018_CVPR}.
    
    The loss function itself is a focus metric (e.g., variance of the IWE pixel intensities) that quantifies the sharpness of the IWE. The network is trained to maximize this focus metric, thereby learning to produce flow fields that best align the events in time. This provides a dense and physically grounded training signal that does not require any ground truth flow labels.

\subsection{Training Setup \label{sec:appendix_train_setup}}
    We train our model on the EVIMO training split~\cite{mitrokhin2019ev} and DSEC training split~\cite{gehrig2021dsec}, aggregating events into 50\,ms windows and encoding them as a two-bin voxel grid. Since EVIMO and DSEC provide Ground Truth at 10\,Hz (i.e., $\Delta t = 100\,\text{ms}$), we supervise the binary-mask loss $\mathcal{L}_{\text{mask}}$ at these timestamps and compute the unsupervised term $\mathcal{L}_{\text{unsup}}$ from the events within the corresponding intervals.

    The loss weights are set to $\lambda_{\text{future\_mask}}=1.0$, $\lambda_{\text{flow}}=0.1$, $\lambda_{\text{smooth}}=0.05$, $w_{\text{sup}}=1.0$, and $w_{\text{unsup}}=1.0$. We did not observe significant performance variations when adjusting these parameters, suggesting that the model is relatively robust to their specific values.

    As EVIMO does not provide ground-truth optical flow, the supervised component of the flow loss is disabled. In contrast, DSEC dataset includes optical flow supervision, but it is only valid in static regions of the scene; therefore, it does not offer a reliable supervisory signal for Independently Moving Objects (IMOs). For this reason the supervised part of the optical flow loss is masked and only the optical flow outside of the IMOs is supervised.

\section{Experiments}

    In this section, we provide a deeper breakdown of the experimental setup.
\subsection{Motion Segmentation on EED \label{sec:EEDtest}}
\begin{table}[t]
    \centering
    \begin{minipage}[t]{0.52\columnwidth}
        \centering
        \scriptsize
        \setlength{\tabcolsep}{2pt}
        \caption{Run time [ms], age, \& GFLOPS}
        \label{tab:run_time}
        \begin{tabular}{ l c c c }
            \toprule
            \textbf{Alg.} & \textbf{Time ($\mu$/$\sigma$)} & \textbf{Age} & \textbf{GFLOPS} \\
            \midrule
            EV-IMO & 8.82/0.54 & -8.8 & 1.80 \\
            \grayrow OMS & 112.7/3.5 & -112.7 & \textbf{0.12} \\
            Ours & \textbf{5.76}/0.64 & \textbf{94.2} & 32.15 \\
            \bottomrule
        \end{tabular}
    \end{minipage}%
    \hfill
    \begin{minipage}[t]{0.46\columnwidth}
        \centering
        \scriptsize
        \setlength{\tabcolsep}{2pt} 
        \caption{IMO pred. (mIoU\%) on EED}
        \label{tab:EED_mIoU}
        \begin{tabular}{ l c c c c c }
            \toprule
            \textbf{Alg.} & \textbf{FD} & \textbf{OC} & \textbf{WIB} & \textbf{LV} & \textbf{MO} \\
            \midrule
            EBMS & 27.2 & 69.7 & 68.4 & 31.0 & 41.6 \\
            \grayrow PMS & 53.9 & \textbf{83.2} & \textbf{93.3} & 48.3 & 51.1 \\
            Ours & \textbf{72.6} & 78.5 & 79.6 & \textbf{63.3} & \textbf{55.4} \\
            \bottomrule
        \end{tabular}
    \end{minipage}
\end{table}
    To evaluate the robustness of our approach to varying illumination conditions and severe sensor noise, we additionally benchmark our method on the Event-based Ego-motion and Dynamic-object (EED) dataset \cite{EED_mitrokhin2018event}. Unlike standard indoor scenes, EED features highly challenging sequences characterized by low illumination and complex, unconstrained trajectories of multiple independently moving objects. 
    
    We compare our method against two recent event-based motion segmentation approaches: EBMS \cite{stoffregen2019event} and ProgressiveMotionSeg (PMS) \cite{chen2022progressivemotionseg}. As shown in Tab.~\ref{tab:EED_mIoU}, performance is measured using the mean Intersection over Union (mIoU \%) across five distinct sequence categories: Fast Drone (FD), OC, WIB, Light Variation (LV), and Multiple Objects (MO).
    
    Our model demonstrates highly competitive performance, achieving state-of-the-art results in the FD, LV, and MO categories. Specifically, our approach outperforms the recent PMS baseline by a significant margin in low-light conditions (LV: 15.0\%) and scenes with multiple dynamic agents (MO: 4.3\%). This underscores the efficacy of our anticipatory suppression mechanism, which successfully isolates IMOs even when standard brightness constancy assumptions fail or event generation is sparse due to darkness. 
    
    While PMS achieves higher accuracy in the OC and WIB sequences, our method maintains strong, consistent performance across the board (outperforming EBMS in all categories). Ultimately, these results prove that our framework generalizes well beyond standard DSEC driving scenarios and EVIMO tabletop scenes, remaining robust in noisy, real-world dynamic environments.

\subsection{Used Metrics\label{sec:appendix_metrics}}
    In Sec.~\ref{sec:futureevimotest} we compute two different metrics to evaluate the models performances: the mean Intersection over Union (mIoU) and the ratio of correctly segmented objects (R).
    The mIoU is computed as the average of the IoU over ego-motion pixels and over independently moving objects (IMOs). Since our model outputs a binary mask (IMO=1, ego-motion=0) rather than per-instance labels, we derive instance matches by computing a one-to-one assignment between predicted and ground-truth instances using the Hungarian algorithm on the IoU cost. An instance is deemed \emph{correct} if, for its matched pair $(\hat{M}_i,M_i)$, $\mathrm{IoU}(\hat{M}_i,M_i) > 0.5$. 
    The ratio of correctly segmented objects with an overlap of more than 50\%
    (R@0.5) is computed as follows:
    \[
    \mathrm{R@0.5} \;=\; \frac{\#\{\text{matched GT with IoU}>0.5\}}{\#\{\text{GT instances}\}}\times 100\%.
    \label{eq:R@0.5}
    \]

\subsection{Time Comparison Details \label{sec:appendix_time_comparison}}
    All timing experiments were conducted on a workstation equipped with a single NVIDIA GeForce RTX 2080 Ti GPU and an Intel Core i9 CPU. To ensure fair comparisons and minimize the impact of multi-threading optimizations in baseline implementations, we restricted the number of usable CPU threads to 1 for all CPU-bound operations. 
    
    We measured the wall-clock inference time by processing a randomized, constant-sized input event tensor. The input randomization prevents caching mechanisms from artificially accelerating repeated inference passes. For each method, we report the mean $\mu$ and standard deviation $\sigma$ over 1000 trials.
    
    \paragraph{GFLOPS Measurement}
        To quantify theoretical complexity, we adopted two distinct strategies based on the nature of the algorithm:
        \begin{itemize}
            \item \textbf{Deep Learning Models (Ours, EV-IMO):} For our method and EV-IMO \cite{mitrokhin2019ev}, which are implemented as standard PyTorch modules, we calculated the floating-point operations using the \texttt{FlopCountAnalysis} tool from the \texttt{fvcore.nn} library. This provides a precise count of the multiply-accumulate operations in the convolutional layers.
            \item \textbf{Algorithmic Baselines (OMS):} Since OMS \cite{clerico2408retina} relies on custom bio-inspired filtering logic rather than standard neural network layers, standard profilers are inapplicable. We therefore estimated its complexity theoretically, following the algorithmic complexity analysis provided in the original paper, summing the per-event operations required for the retina-inspired center-surround filtering.
        \end{itemize}
        For all the algorithms we consider the same input size of $480 \times 640 \times 2$.
    
    \paragraph{Definition of Prediction Age}
        Standard latency metrics often overlook the "freshness" of the output in robotic control loops. We define \textit{Prediction Age} ($T_{age}$) as the temporal offset between the moment a mask becomes available to the system and the timestamp it describes. It is calculated as: $T_{age} = \Delta t_{forecast} - T_{runtime}$
        where $\Delta t_{forecast}$ is the prediction horizon (0 for baselines, 100\,ms for our method) and $T_{runtime}$ is the wall-clock inference time.
        \begin{itemize}
            \item \textbf{Negative Age ($T_{age} < 0$):} Indicates \textit{system lag}. The mask describes a scene that happened $T_{age}$ milliseconds ago. Both EV-IMO ($T_{age} \approx -9$\,ms) and OMS ($T_{age} \approx -113$\,ms) fall into this category.
            \item \textbf{Positive Age ($T_{age} > 0$):} Indicates \textit{anticipation}. The mask describes a future scene state, remaining valid even after the computation time elapses. Our method achieves $T_{age} = 100 - 5.8 = 94.2$\,ms, providing a "zero-latency" output effectively.
        \end{itemize}

    \begin{figure*}[t]
        \centering
        \includegraphics[width=\textwidth]{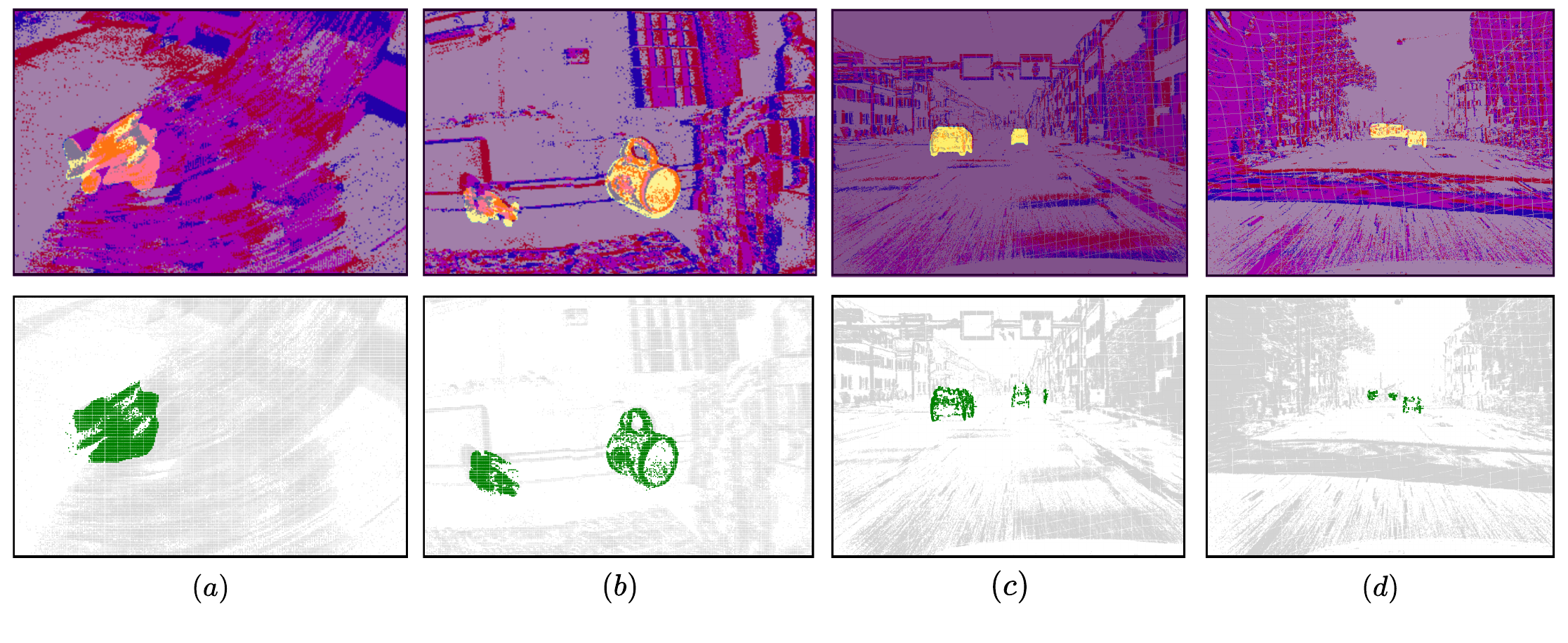}
        \caption{Visualization of anticipatory motion suppression on EVIMO (a-b) and DSEC (c-d). Top row: overlaid events accumulated over $\Delta t_p = 100\,\text{ms}$ with ground-truth masks for independently moving objects (IMO). Bottom row: events attributed to ego-motion (white) and to IMO (green), obtained by applying a motion mask to future $\Delta t_p = 100\,\text{ms}$ events, with motion predicted from the previous $\Delta t = 50\,\text{ms}$ events using our Anticipatory Motion Suppression pipeline. Panels: (a) Robustness under fast ego-motion and non-convex masks, recovering IMO events from a large stream of ego-motion events. (b) Performance in indoor, multi-object scenes, with arbitrary shaped non-convex objects. (c-d) Autonomous driving results, anticipating the motion of small moving objects, both far-away vehicles and nearby objects.}
        \label{fig:model_results_appendix}
    \end{figure*}

\subsection{Qualitative Results\label{sec:appendix_qualitative}}
    \subsubsection{Motion Segmentation on EVIMO and DSEC}
    In addition to the primary qualitative results presented in the main text, we provide further visualizations in Fig.~\ref{fig:model_results_appendix} demonstrating our method's robustness across diverse indoor and outdoor scenarios.
    
    \textbf{Complex Geometries and Clutter (EVIMO):}
        Panel (b) of Fig.~\ref{fig:model_results_appendix} illustrates the performance on a cluttered indoor scene featuring multiple arbitrary objects. Unlike standard bounding-box approaches, our pixel-wise segmentation accurately captures the non-convex geometry of the mug (e.g., the handle hole) and the complex contours of the toy figure. The bottom row shows that even with significant background clutter, the anticipatory mask successfully isolates the exact pixels of the independently moving objects (green) while suppressing the dense background ego-motion events (white).
    
    \textbf{Scale Variance in Driving Scenes (DSEC):}
        Panels (c-d) highlight the system's capability in dynamic automotive environments. The scene contains multiple vehicles moving at different relative speeds and distances. Our method correctly anticipates and segments both the nearby vehicle (c) and the smaller, distant vehicles further down the road (d). Our model leverages strong motion priors from EVIMO and DSEC to segment dynamic objects not explicitly labeled in the ground truth, such as pedestrians. In panel (c), for example, the system correctly identifies a pedestrian (third from left) as an IMO, highlighting its capacity for motion segmentation beyond the ground-truth labeled classes.
        
    \subsubsection{Motion Segmentation in Challenging Scenarios}
    \begin{figure*}[t]
    \centering
    \begin{minipage}[t]{0.5\columnwidth}
        \centering
        \includegraphics[width=\linewidth]{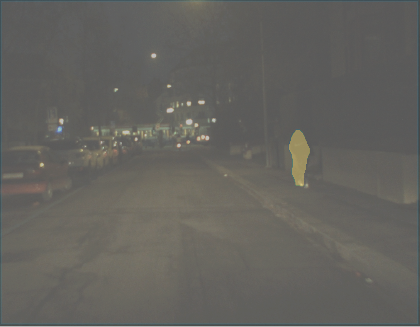}
    \end{minipage}\hfill
    \begin{minipage}[t]{0.5\columnwidth}
        \centering
        \includegraphics[width=\linewidth]{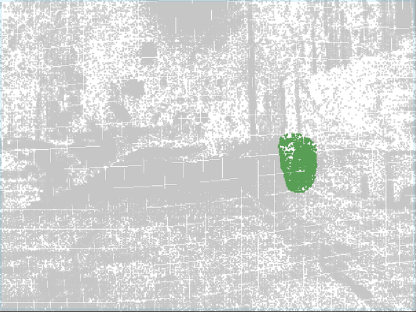}
    \end{minipage}\hfill
    \begin{minipage}[t]{0.5\columnwidth}
        \centering
        \includegraphics[width=\linewidth]{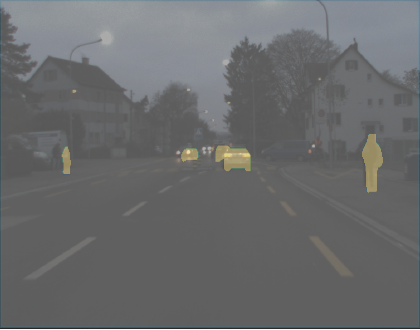}
    \end{minipage}\hfill
    \begin{minipage}[t]{0.5\columnwidth}
        \centering
        \includegraphics[width=\linewidth]{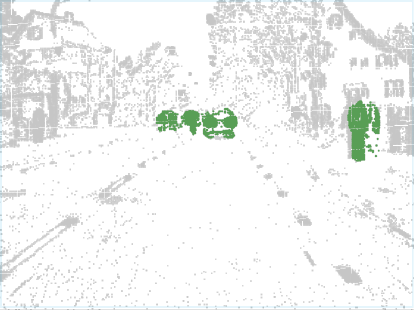}
    \end{minipage}


    \caption{Qualitative results on DSEC under challenging lighting conditions. (a, b) In night scenes, our method isolates dynamic objects despite sparse events and sensor noise, though flickering headlights can slightly inflate pedestrian boundaries. (c, d) In low-contrast cloudy scenarios, vehicles are reliably segmented across various distances. However, objects generating minimal temporal contrast (e.g., the missing pedestrian on the far left) may lack sufficient motion cues for detection. Yellow masks indicate ground truth; green indicates our predictions.}
    \label{fig:dsec_low_light}

\end{figure*}
    
\begin{figure*}[t]
    \centering
    \begin{minipage}[t]{0.5\columnwidth}
        \centering
        \includegraphics[width=\linewidth]{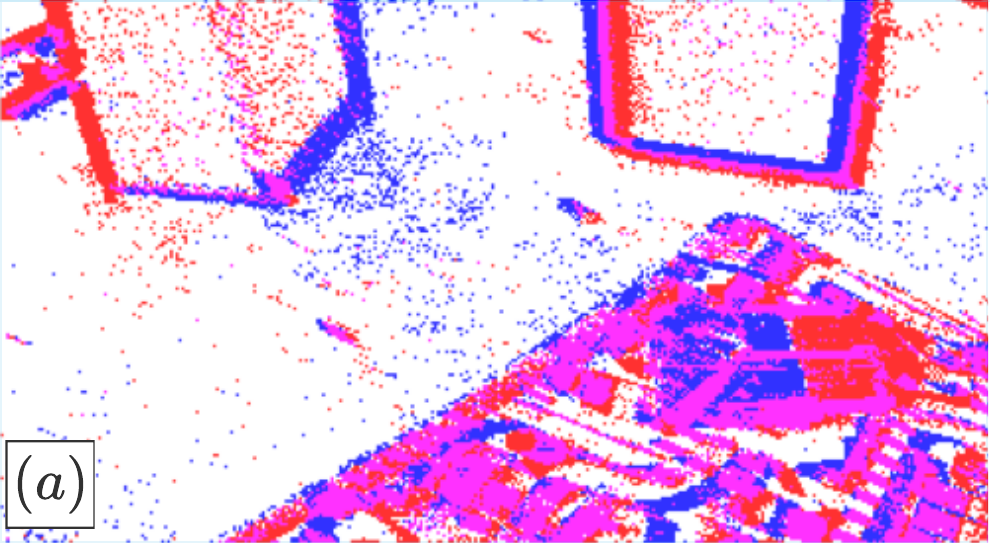}
    \end{minipage}\hfill
    \begin{minipage}[t]{0.5\columnwidth}
        \centering
        \includegraphics[width=\linewidth]{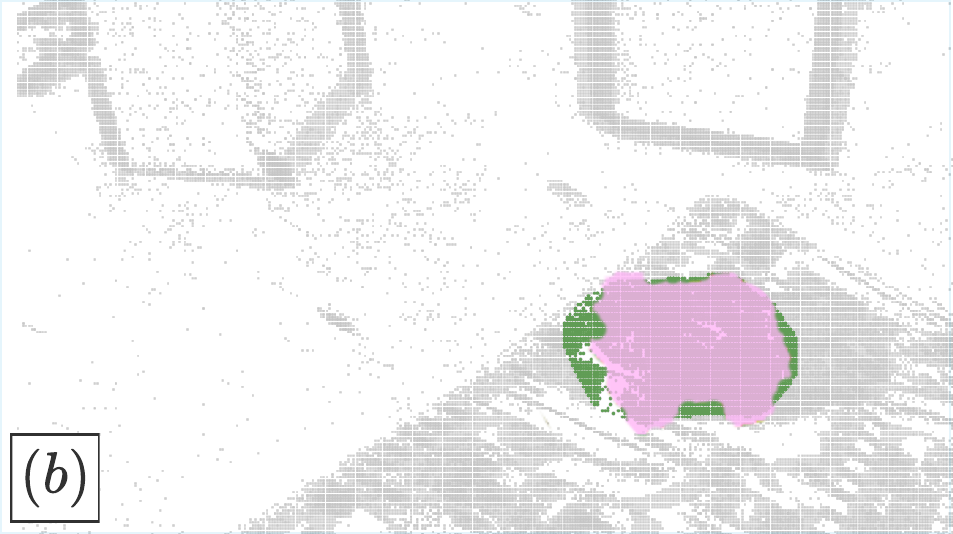}
    \end{minipage}\hfill
    \begin{minipage}[t]{0.5\columnwidth}
        \centering
        \includegraphics[width=\linewidth]{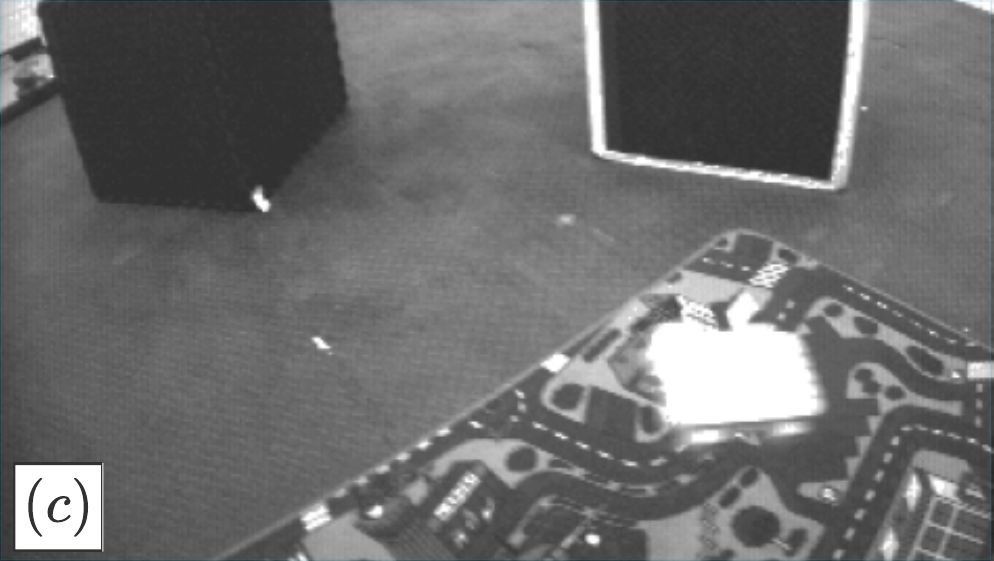}
    \end{minipage}\hfill
    \begin{minipage}[t]{0.5\columnwidth}
        \centering
        \includegraphics[width=\linewidth]{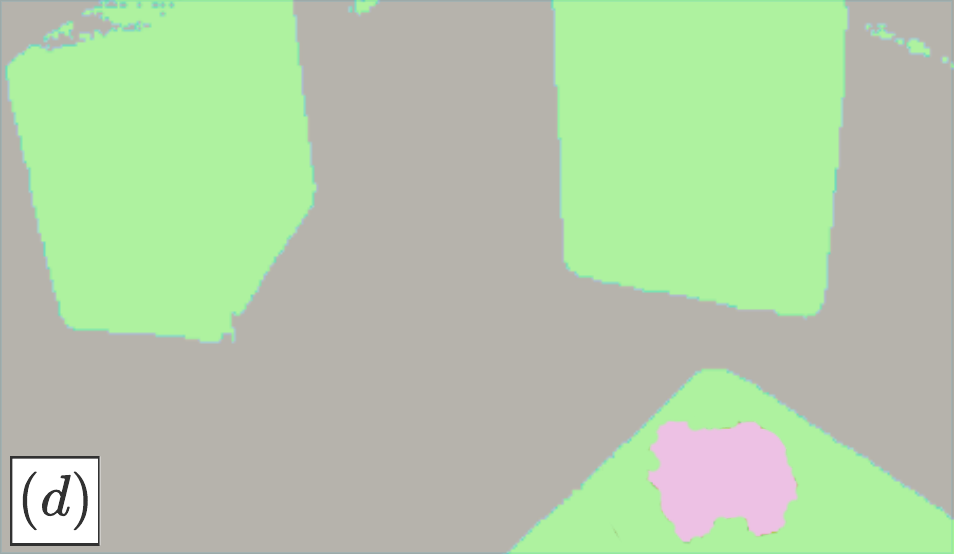}
    \end{minipage}

    \caption{(a) EVIMO events. (b) Our IMO prediction (green) vs. ground truth (purple). (c) Overexposed frame. (d) FlowSAM prediction from (c), green: prediction, purple: ground truth.}
    \label{fig:evimo_flowsam}

\end{figure*}
    Event cameras natively possess a high dynamic range, allowing them to capture scene dynamics even in visually degraded conditions where standard frame-based cameras often fail. We demonstrate our method's robustness on the DSEC dataset under challenging lighting conditions reporting qualitative results in Fig.~\ref{fig:dsec_low_light}.

     \textbf{Night-time and Low-Light Driving:} In the night scenes shown in Fig.~\ref{fig:dsec_low_light}a and \ref{fig:dsec_low_light}b, standard brightness constancy assumptions frequently break down, leading to event streams that are either sparse or degraded by sensor noise. Despite these extreme conditions, our anticipatory suppression mechanism successfully isolates vehicles and pedestrians in profound darkness. While the predicted dynamic masks (green) closely align with the ground truth (yellow), we note a minor limitation: the high event rate triggered by flickering car headlights causes the model to overpredict the pedestrian boundaries. Consequently, the predicted mask becomes larger than the ground truth, introducing localized false positives.

     \textbf{Cloudy and Flat-Light Conditions:} Similarly, in cloudy or flat-lighting scenarios where visual contrast is low, our method reliably segments multiple moving vehicles from the background ego-motion at various scales and distances. While the network generally groups sparse events into cohesive instance masks despite the lack of distinct visual edges, it can struggle with objects that produce minimal temporal contrast. For instance, a pedestrian on the left side of the scene is missing from our prediction; because this target generates an extremely sparse event stream, it fails to provide sufficient motion cues to be successfully disentangled from the background.

     \subsubsection{Qualitative Comparison with Frame-Based Methods (FlowSAM)}

     To further illustrate the fundamental advantages of event-based motion segmentation over traditional frame-based methods, we present a qualitative comparison against FlowSAM in Fig.~\ref{fig:evimo_flowsam}. To ensure a strong baseline, we implement FlowSAM by supplying it with dense, high-quality optical flow estimates generated by RAFT, which serve as motion prompts to guide its Segment Anything (SAM) backbone. However, standard RGB cameras are inherently limited by their dynamic range and exposure times, making them highly susceptible to motion blur and overexposure in dynamic environments. As seen in Fig.~\ref{fig:evimo_flowsam}.c, rapid movement and complex lighting conditions cause the object to become severely overexposed, stripping away vital visual features and distinct spatial edges. Consequently, even when aided by RAFT-estimated optical flow, state-of-the-art image-based models like FlowSAM fail to accurately segment the independently moving object, as shown in Fig.~\ref{fig:evimo_flowsam}.d where the predicted mask (green) drastically deviates from the ground truth (purple). 

    Conversely, event cameras respond strictly to asynchronous brightness changes, natively bypassing sensor saturation and motion blur. As demonstrated in Fig.~\ref{fig:evimo_flowsam}.a, the event stream preserves the precise, high-speed contours of the moving object. By leveraging this high-fidelity temporal data, our anticipatory suppression pipeline successfully isolates the IMO mask (Fig.~\ref{fig:evimo_flowsam}.b), tightly aligning our prediction (green) with the ground truth (purple) even when traditional visual data is entirely corrupted.

\subsection{Sensitivity Study}

\subsubsection{Sensitivity to Input Event Rate\label{sec:appendix_sensitivity_rate}}

    \begin{figure}[t]
        \centering
        \includegraphics[width=\columnwidth]{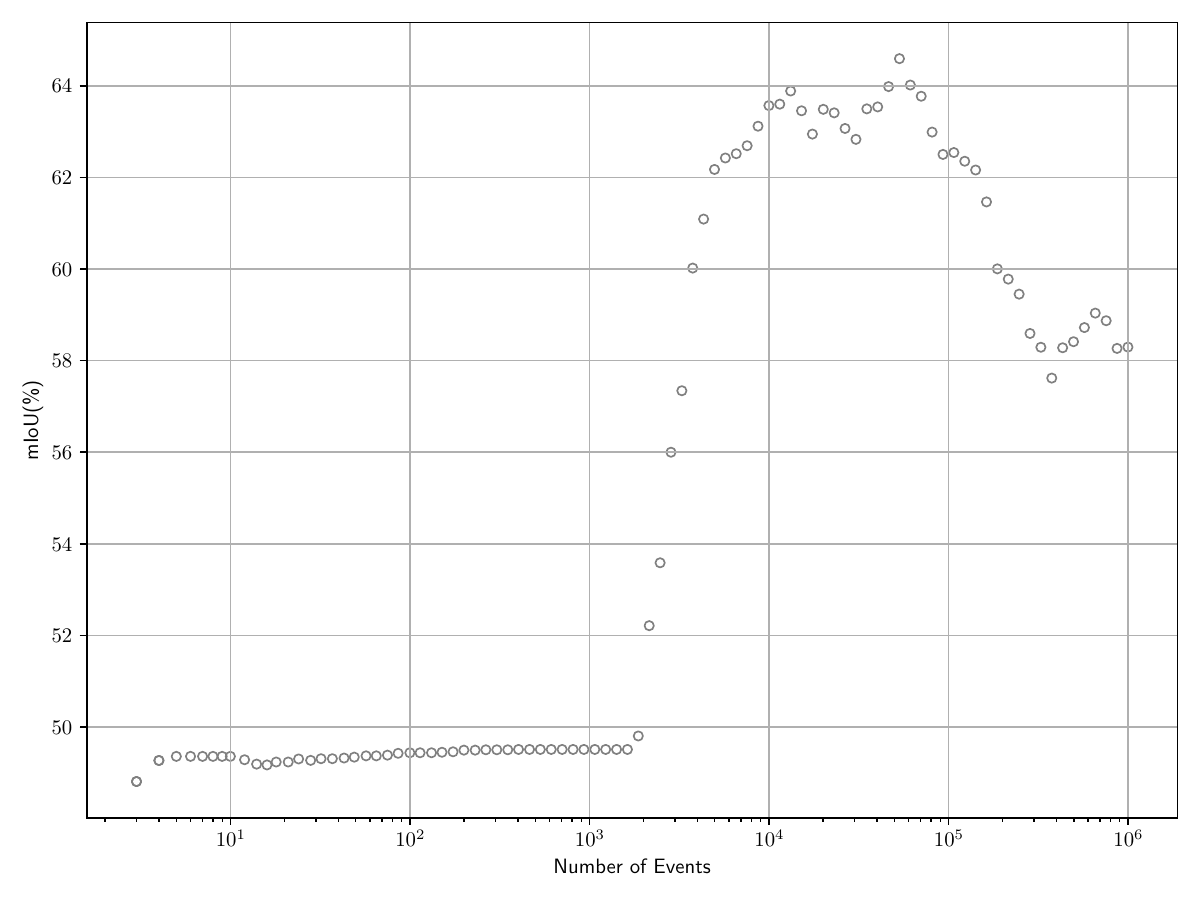}
        \captionof{figure}{%
        Segmentation accuracy (mIoU) as a function of the number of input events $N$ used to form the network input. The x-axis is logarithmic. At test time, instead of fixing a temporal window $\Delta t$, we fix an event budget $N$ and measure performance. Accuracy is poor for very sparse inputs ($N < 10^3$), improves steadily, and reaches its maximum around $N \approx 6\times 10^4$. Beyond this point, performance gradually declines as overly long temporal integration blurs object boundaries. The steady region ($10^4 \!-\!10^5$ events) indicates a broad operating plateau in which the method is both accurate and stable across different scene dynamics.}
        \label{fig:sensitiviy}
    \end{figure}
    
    Event cameras emit asynchronous brightness changes at a rate that depends on scene texture, ego-motion, lighting, and independently moving objects (IMOs). As a consequence, a fixed wall-clock integration window $\Delta t$ does not correspond to a fixed number of events across different operating conditions. Fast motion, rich texture, or high contrast may yield tens of thousands of events in tens of milliseconds, whereas slow motion in low-texture regions may produce only a few hundred events over the same $\Delta t$. In other words, the effective input density seen by the network is governed primarily by the number of events $N$, not by elapsed time. For this reason, we study segmentation accuracy as a function of $N$ rather than as a function of $\Delta t$.
    
    For this analysis, we used the model trained on the EVIMO dataset~\cite{mitrokhin2019ev}. Training follows our standard setup as described in Sec.\ref{sec:train_setup}. At evaluation time, however, we depart from the fixed $\Delta t$ input: instead of always feeding the last 100\,ms of events, we construct inputs by taking a \emph{fixed event budget} of $N$ events immediately preceding each ground-truth timestamp and discarding the rest.
    
    We evaluate this procedure on a subset of the EVIMO test set. For each value of $N$, we compute the mean Intersection over Union (mIoU) averaged across all samples in all trajectories.

    Figure~\ref{fig:sensitiviy} reports mIoU versus $N$ (log scale on the x-axis). The curve is non-monotonic and reveals three distinct regimes:
    \begin{itemize}
        \item \textbf{Sparse regime ($N < 10^3$ events).} With fewer than $\sim10^3$ events, segmentation accuracy is low (mIoU $\approx 49$--$50\%$). In this regime, the network receives too little contrast to reliably segment ego-motion from IMO motion. On EVIMO, the camera is almost always moving, and IMOs typically occupy only a small fraction of the field of view. When so few events are available, most activity in the aggregated event frame appears dominated by ego-motion, and the model tends to over-label events as background. The result is that the static background is often segmented consistently, but independently moving objects are systematically under-segmented.
        \item \textbf{Plateau / optimal regime ($10^4$--$10^5$ events).} As $N$ increases, performance improves fast and peaks around $N \approx 6 \times 10^4$ events. Importantly, there is a broad plateau between roughly $10^4$ and $10^5$ events in which mIoU remains high and stable. This shows that the method does not need a finely tuned event budget: once there are enough events to capture both ego-motion and IMO motion, it works reliably across a wide range of scene dynamics and event rates. At deployment, this plateau defines a practical operating range for asynchronous processing: triggering inference whenever $\sim10^4$--$10^5$ new events have accumulated yields near-optimal segmentation quality.
        \item \textbf{Dense regime ($N \gtrsim 10^5$ events).} Beyond the optimal plateau, adding more events decreases mIoU. Very large $N$ implicitly integrates over a much longer temporal span $\Delta t$. Over such long horizons, independently moving objects drift noticeably, so their boundaries appear blurry in the accumulated representation, which reduces the IoU.
    \end{itemize}
    
    Overall, this experiment shows that the model’s accuracy depends on event density in a predictable way: too few events hurt because there is not enough motion evidence, too many hurt because boundaries blur. Between $10^4$ and $10^5$ events there is a broad, stable high-performance plateau, showing robustness to changing scene dynamics.

\subsubsection{Sensitivity to Object Velocity\label{sec:appendix_sensitivity_velocity}}
\begin{table}[t]
    \centering
    \footnotesize 
    \caption{EVIMO (mIoU\%) by Speed (px/s)}
    \label{tab:evimo_by_dynamics}
    \begin{tabular}{ l c c c }
        \toprule
        \textbf{Alg.} & \textbf{$\leq$90} & \textbf{90-120} & \textbf{$\geq$120} \\
        \midrule
        EV-IMO [10] & 77.8 & 76.9 & 66.9 \\
        \grayrow Ours & \textbf{82.3} & \textbf{79.7} & \textbf{72.2} \\
        \bottomrule
    \end{tabular}
    \label{tab:sensitivity:obj_speed}
\end{table}
To further analyze how scene dynamics influence segmentation performance, we evaluate our method across different object velocity regimes. We partition the EVIMO dataset into three categories based on the relative speed of the independently moving objects (IMOs) on the image plane, measured in pixels per second (px/s): slow ($\leq 90$), medium ($90-120$), and fast ($\geq 120$). 

The mean Intersection over Union (mIoU \%) for each partition is presented in Table~\ref{tab:sensitivity:obj_speed}. As expected, there is a general monotonic decrease in accuracy for both our approach and the baseline EV-IMO [10] as object speed increases. Faster motion inherently increases the difficulty of the segmentation task due to larger inter-frame displacements, more severe occlusions, and the exacerbation of prediction-age latency (where even a few milliseconds of delay translates to several pixels of spatial misalignment). 

Despite these challenges, our proposed anticipatory motion suppression framework consistently outperforms EV-IMO across all velocity profiles. Notably, even in the most challenging highly dynamic regime ($\geq 120$ px/s), our method maintains a strong mIoU of $72.2\%$ (a $+5.3$ point improvement over EV-IMO). This demonstrates that our learned flow forecasting effectively mitigates the latency and spatial misalignment issues that typically degrade performance in fast-moving scenarios, confirming the robustness of our architecture to diverse motion regimes.

\end{document}